\pdfoutput=1

\documentclass[11pt]{article}

\usepackage{EMNLP2023}
\usepackage{times}
\usepackage{multirow}
\usepackage{latexsym}
\usepackage{graphicx}
\usepackage{hyperref}

\usepackage[T1]{fontenc}

\usepackage[utf8]{inputenc}
\usepackage{float} 
\usepackage{microtype}

\usepackage{inconsolata}

\usepackage{xcolor}

\usepackage{booktabs}

\usepackage{soul}

\title{Using Perspectival Words Is Harder Than Vocabulary Words for Humans\\—and Even More So for Multimodal Language Models}

\renewcommand{\thefootnote}{\fnsymbol{footnote}}
\author{
Dota Tianai Dong\thanks{\hspace{0.5em}Equal contribution}\quad
Yifan Luo\footnotemark[1]\hspace{0.3em}\thanks{\hspace{0.5em}Work has been done as an student assistant at MPI}\quad
Po-Ya Angela Wang \quad
Asli Özyürek \quad
Paula Rubio-Fernández\\
Max Planck Institute for Psycholinguistics \\
\texttt{\{tianai.dong, yifan.luo, amber.wang, asli.Ozyurek, paula.rubiofernandez\}@mpi.nl}
}

\begin{document}
\maketitle
\begingroup
\renewcommand{\thefootnote}{} 
\footnotetext{Codes available at \url{https://github.com/Beckinetic/VLMIndexical}}
\endgroup

\begin{abstract}
Multimodal language models (MLMs) increasingly demonstrate human-like communication, yet their use of everyday perspectival words remains poorly understood. To address this gap, we compare humans and MLMs in their use of three word types, which we predict impose increasing cognitive demands: vocabulary (e.g., `boat' or `cup'), possessives (e.g., `mine' vs. `yours'), and demonstratives (e.g., `this one' vs. `that one'). Testing seven MLMs against human participants, we find that perspectival words are harder than vocabulary words for both groups. The gap is even larger for MLMs: while models approach human-level performance on using vocabulary, they exhibit clear deficits with possessives and even greater difficulties with demonstratives. Ablation analyses point to limitations in perspective-taking and spatial reasoning as key sources of these gaps in MLMs. Instruction-based prompting helps close the gap for possessives but still leaves demonstratives far below human performance. These results show that, unlike vocabulary, perspectival words pose a greater challenge in human communication—and this difficulty is further amplified in MLMs, revealing a crucial shortfall in their pragmatic and social-cognitive abilities.
\end{abstract}
\section{Introduction}
A central goal in NLP has long been to develop systems that can communicate about the world as humans do. Recent advances in multimodal language models (MLMs) have brought us closer to this goal, enabling systems to use text and image so naturally that users often perceive them as real conversational partners \citep{elkins2020can}. Existing evaluations of MLMs have largely focused on vocabulary—words with relatively stable, context-independent meanings in conversation, such as object names, colors, and verbs \citep{yuksekgonul2022and,chaoyou2023mme}. This vocabulary-centric approach, however, overlooks a fundamentally different yet ubiquitous word class: perspectival words such as `I', `you', `mine', and `this one'.\footnote{`Perspectival words' is a non-technical label for indexicals, specifically personal, possessive, and demonstrative pronouns.}

Unlike vocabulary words, perspectival words are part of grammar, and their meaning is \textbf{grounded in the context of use}. For example, `I' and `you' refer to the speaker and listener in a conversation, so they denote a different interlocutor at every turn. Perspectival words are also sensitive to \textbf{speaker-listener perspectives}: whether a given spoon is `mine' or `yours' depends on who is speaking, and whether it counts as `this one' or `that one' depends on the object's spatial position relative to the speaker's \citep{rubio2021pragmatic,rubio2022demonstrative}. Therefore, using perspectival words requires \textbf{pragmatics}---the ability to ground language use in its physical and social context, and \textbf{social cognition} (or `Theory of Mind')---the ability to understand others' minds and take their perspective.  

\begin{figure}[!t]
\begin{center}
\includegraphics[width=1\columnwidth]{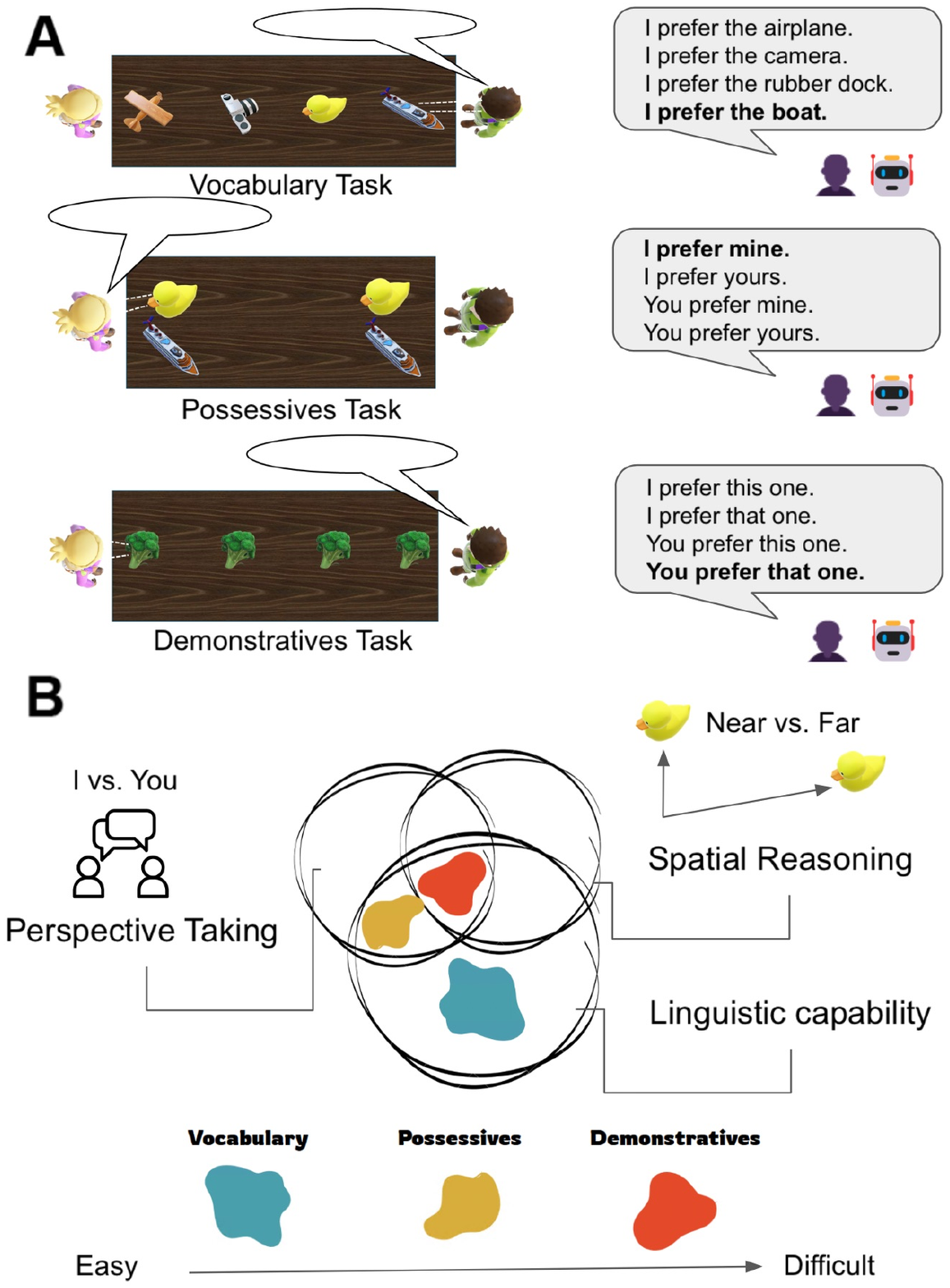}
\end{center}
\caption{\textbf{A:} We test human participants and MLMs across three tasks: (1) Vocabulary task: selecting the correct label for a referent (e.g., `boat'); (2) Possessive task: selecting `I' vs `you' and `mine' vs `yours', and (3) Demonstrative task: selecting `I' vs `you' and `this one' vs `that one'. \textbf{B:} We predict a hierarchy of task difficulty.}
\label{figure-1}
\end{figure}


Despite their frequent and central role in everyday communication \citep{clark1996using,carston2002thoughts,levinson2006deixis,levinson2024dark,rubio2021pragmatic,rubio2024cultural,jara2024demonstratives}, perspectival words have received surprisingly little attention in MLM evaluation. This gap is consequential because correctly using these words requires pragmatic reasoning and perspective-taking. Investigating how MLMs use perspectival words therefore provides a direct test of whether these systems possess the pragmatic capabilities \citep{jeretic2020natural,tong2021recent,liu2022testing,ruis2022large,stowe2022impli,barattieri2023pragmatic,hu-etal-2023-fine} and Theory of Mind \citep[for reviews, see][]{ma2023towards,shapira2023clever,hu2025re} that are necessary for genuinely human-like communication.

In cognitive science, perspectival words are also less well understood than sentence-level pragmatic phenomena (e.g., metaphor, irony, or scalar implicature). Recent behavioral and computational work suggests that perspectival words rely on \textbf{social micro-processes} \citep{rubio2024tracking}:  temporally-bound mentalistic representations that allow us to reason about others during real-time communication, without fully modeling their minds \citep[see also][]{holler2025facial}. Comparing how humans and models use perspectival words can therefore offer valuable insights into the socio-cognitive basis of conversation---for instance, whether pragmatic behaviors can emerge without explicit mentalistic representations \citep{andreas2022language}, or in the absence of communicative interaction in the physical world \citep{fried2023pragmatics}.

We introduce a multiple-choice task to test two novel hypotheses in both humans and MLMs. First, selecting the correct label for an object (e.g., `spoon' vs. `boat') should be easier than using possessives (`mine' vs. `yours') or demonstratives (`this one' vs. `that one') for the same objects (Figure \ref{figure-1}). Second, we predict that demonstrative use will be more challenging than possessive use, as it requires both perspective-taking and spatial reasoning. To evaluate these hypotheses, we collected data from 360 adult participants using materials carefully designed to avoid training contamination in MLMs \citep{akata2023playing}, and compared human performance with that of seven state-of-the-art MLMs.

We observe the predicted hierarchy of word usage difficulty in both humans and most MLMs: Vocabulary < Possessives < Demonstratives. For Vocabulary, models achieve near-human ceiling performance, but their accuracy drops below human levels for possessives and declines even further for demonstratives. Ablation analyses further suggest these difficulties arise from limitations in perspective-taking and spatial reasoning in the models. Finally, instruction-based prompting improves model performance with possessives, bringing it close to human-level accuracy, while gains on demonstratives remain far below human performance, and primarily reflect enhanced perspective-taking rather than improved demonstrative use. Overall, these findings provide novel insights into the cognitive demands of perspectival words relative to vocabulary words for both humans and MLMs, and highlight the challenge of producing grammatical forms that require pragmatic reasoning and social cognition—capacities that remain limited in current NLP systems.

\section{Related work}
Prior research on NLP systems' pragmatic abilities has mostly focused on non-literal language use. \citet{hu-etal-2023-fine}, for example, tested whether humans and models select pragmatic interpretations of utterances representing seven pragmatic phenomena (i.e., deceits, indirect speech, irony, conversational maxims, metaphor, humor and coherence). \citet{sravanthi2024pub} released a Pragmatics Understanding Benchmark (PUB) including fourteen tests of four pragmatic phenomena (implicature, presupposition, metonymy and co-reference) using multiple-choice question answers. Findings from these studies suggest that language models demonstrate pragmatic language comprehension comparable to humans across multiple tasks. However, with the exception of co-reference \citep[see also][]{zheng2021grice}, all evaluated pragmatic phenomena are non-literal language uses, which limits pragmatics to the message level \citep[cf.][]{carston2002thoughts}. Work on perspectival aspects of meaning remains very limited. One exception is \citet{masis-anderson-2021-prosper}, who examine perspectival motion verbs in language models and show that models represent spatial perspective differently from humans. Yet, this work is limited to textual settings and does not include multimodal interaction or face-to-face reference. As a result, these model evaluations rest on a skewed view of pragmatics, casting doubts on models' communicative abilities once word-level pragmatics and face-to-face interaction are considered \citep{bisk2020experience,rubio2024tracking}. Our work differs from these approaches in two key respects: by testing physical reference, we study word-level pragmatics and ground language use in a simulation of face-to-face interaction. 

Using perspectival words requires perspective-taking, which is part of human social cognition. Language models have been extensively evaluated for their Theory of Mind skills, with mixed results \citep[for review and discussion, see][]{hu2025re}. 
In human communication, the role of Theory of Mind has traditionally been limited to the sentence level (e.g., interpreting `What a great friend' as a sarcastic remark). However, it has recently been argued that human communication also involves social micro-processes \citep{rubio2024tracking}. As shown by behavioral and computational work \citep{rubio2021speakers,jara2022social,jara2024demonstratives,woensdregt2022language}, physical reference relies on social micro-processes. Our study is the first to compare humans and MLMs on this form of social cognition.

Our work is also relevant to a substantial body of research on spatial language in multimodal models. Recent vision-language benchmarks have examined models’ ability to interpret relative spatial expressions such as `left' and `right,' often revealing systematic weaknesses in spatial reasoning \citep{kamath-etal-2023-whats,liu-etal-2023-visual}. However, these studies primarily target absolute or relative spatial relations, rather than perspectival words that depend on the speaker's viewpoint. By contrast, we focus specifically on perspectival referential terms, which require perspective-taking and are central to situated, interactive communication.

More broadly, our work connects to research that develops MLMs to align with human dialogue in reference tasks, where participants share access to a set of referents and the speaker describes a target for the listener to identify \citep{de2017guesswhat,das2017visual}. Prior studies show that models often rely on fewer or less effective referential expressions, but that incorporating human-like referential strategies in these contexts improves both task success and conversational quality \citep{shekhar-etal-2018-ask,dong2021visually,testoni2024racquet}. We extend this line of research by introducing a controlled and interpretable framework for investigating how humans and models use perspectival words in simulated face-to-face interaction, providing a cognitive foundation for understanding such referential expressions in naturalistic communication.

\section{Method}
\subsection{Tested Phenomena}
Our experiment investigates word usage in humans and MLMs across three tasks: Vocabulary, Possessive, and Demonstrative tasks, which we hypothesize increase in production difficulty. 

\paragraph{Vocabulary} We predict that both humans and models will be able to identify the referent object and select the correct label. This is the easiest task because it does not require pragmatic grounding or perspective taking \cite{clerkin2022real}.
\paragraph{Possessive pronouns} 
We predict humans will perform better than models, and both groups will perform worse than in the Vocabulary task. This task is harder because it requires pragmatic grounding (i.e., anchoring on the speaker) and perspective taking (i.e., monitoring whose preference is being expressed) \cite{levinson2006deixis}.
\paragraph{Demonstrative pronouns} 
We predict humans will perform better than models, and both groups will perform worse than in the Possessive task. This is the hardest task because it requires not only pragmatic grounding and perspective taking, but also spatial reasoning (i.e., which objects are near vs far from the speaker) \cite{rubio2022demonstrative}.

\subsection{Task Designs}
Each experiment uses a multiple-choice format with four options, only one of which is correct (as shown in Figure \ref{figure-1}). The experimental design systematically manipulates two independent variables: spatial arrangement (i.e., speaker-target distance) and perspective (`I prefer' vs `You prefer'). We selected 128 objects that are reliably recognizable across models and designed two parallel task setups—one optimized for humans and one for MLMs—that present identical communicative scenarios but differ in how information is provided. This design choice is deliberate: by allowing each participant group to process information in their native modality, we preserve their ability to produce the correct referential expression while avoiding confounds (e.g., failure in object identification). We validated this dual design through systematic sanity checks that confirmed models perform better in their own tasks than in those for humans (see Appendix \ref{app:sanity}).

\paragraph{Task setups for humans} We built three separate tasks, with the same instructions: participants would see a series of visual scenes depicting two characters (a boy and a girl) standing on opposite sides of a table with four objects. Participants were asked to play the role of the speaker (who had a speech bubble) and select the correct message from among four options. The speaker was expressing either their own or the other character's preference for one of the four objects (as indicated by their line of gaze). 

In the Vocabulary task, four different objects were placed along the table. In the Possessive task, two identical pairs of objects were placed at either end of the table, with object-character proximity indicating ownership. In the Demonstrative task, four identical objects were placed along the table. Four versions of each visual display were generated, in a fully-crossed Speaker (Boy vs Girl) by Perspective (`I prefer' vs `You prefer) design (see Fig.\ref{figure-1}). Object ownership (in the Possessive task) and distance to the speaker (closest vs furthest, in the Demonstrative task) were fully counterbalanced. In the Vocabulary task, all four possible responses revealed the same perspective (either `I prefer...' or `You prefer...') and each contained the name of one of the objects on the table (e.g., `the plane,' `the camera,' `the rubber duck,' `the boat'). In the Possessive and Demonstrative tasks, two responses expressed the speaker's perspective and another two the other agent's, each followed by either `mine' vs `yours,' or `this one' vs `that one,' in a fully-crossed design.

\paragraph{Task setups for MLMs} 
We designed three tasks for MLMs, each comprising 6144 trials combining an image and a text prompt. Task images mirrored human stimuli but were adapted for model input: speech bubbles and gaze lines were removed, a green square was added to mark the target object, and speaker and gazer identity were included in the prompt. To ensure comparability with human tasks, we applied controlled augmentations to task-irrelevant features (e.g., object order; see Appendix \ref{app:augmentation}). We also designed the stimuli to avoid frame-of-reference ambiguity by explicitly specifying the speaker and gazer, with both image and text conveying clear binary near/far relations to the speaker (see Appendix \ref{app:ExternalFoR}). We designed prompts that closely match human instructions (Appendix \ref{zero-shot}). All models were tested in zero-shot settings with temperature set to zero to evaluate the knowledge learned from training. To ensure that performance truly reflects referential abilities rather than prompt-specific artifacts, we tested multiple prompt variants and confirmed that results remained consistent across formulations (see Appendix \ref{app:prompt variation} for variant comparisons). 

\subsection{Evaluating humans}

\paragraph{Production tasks} The Vocabulary task included 16 trials, and the Possessive and Demonstrative tasks included 32 trials each. The tasks were built using Qualtrics and administered to three separate groups of monolingual native English speakers recruited through Prolific (\textit{N}=120 per task). To ensure that the original results replicated when the experiment was run within participants, a single task was built using 16 trials from each condition distributed in two lists of trial blocks (Vocabulary-Possessives-Demonstratives and Vocabulary-Demonstratives-Possessives) and administered to a different group of native English speakers (\textit{N}=100) in a follow-up task (for details and results, see Appendix \ref{app:humans}).


\subsection{Evaluating MLMs}

\paragraph{MLMs} 
We evaluated seven models (more details in Appendix \ref{app:specs}): (1) GPT-4o, a closed-source model with real-time multimodal reasoning capabilities \cite{openai2024gpt4o}; (2) Llama-4-Scout-17B-16E-Instruct (hereafter Llama-4-Scout), an open-source vision-language model integrating a visual encoder and language model \cite{meta2025llama4}; (3) Gemma-3-27B-it (Gemma-3), which tokenizes images for multimodal input \cite{gemma3report2025}; (4–5) InternVL 2.5-4B-MPO (InternVL 2.5-4B) and 8B-MPO (InternVL 2.5-8B), differing in scale, using a visual encoder and LLM bridged by an MLP projector \cite{chen2024expanding}; and (6–7) Ovis 2-4B and 8B, aligning visual and textual embeddings for enhanced reasoning \cite{lu2024ovis}. We used the LLM-as-a-Judge method \cite{zheng2023judging}, with GPT-4o as the evaluator, and our thorough manual validation of response subsets confirmed complete reliability.

\paragraph{Ablation analysis} To further examine the contribution of spatial and perspectival cues to reference production, we conducted an ablation analysis on the MLMs. Spatial cues were manipulated under three conditions: (i) \textit{Spatial Cue Removal (Linguistic)}: ownership markers (for possessives) and proximity terms (for demonstratives) were omitted from the prompts; (ii) \textit{Spatial Cue Removal (Visual)}: all objects were displayed using identical green squares; and (iii) \textit{Spatial Cue Removal (Visual + Linguistic)}: both linguistic and visual cues were removed, such that ownership markers, proximity terms, and green squares were all removed. For perspectival cues, we implemented (iiii) \textit{Perspectival Cue Removal (Linguistic)}: the speaker’s identity and the gazer’s identity were specified, while removing any explicit indication of the viewpoint to be adopted by the model. Further implementation details are provided in Appendix~\ref{app:ablation}. Addtional baselines in Appendix~\ref{addtionalbaselines}.

\begin{figure*}[t]
\centering
\includegraphics[width=0.85\textwidth]{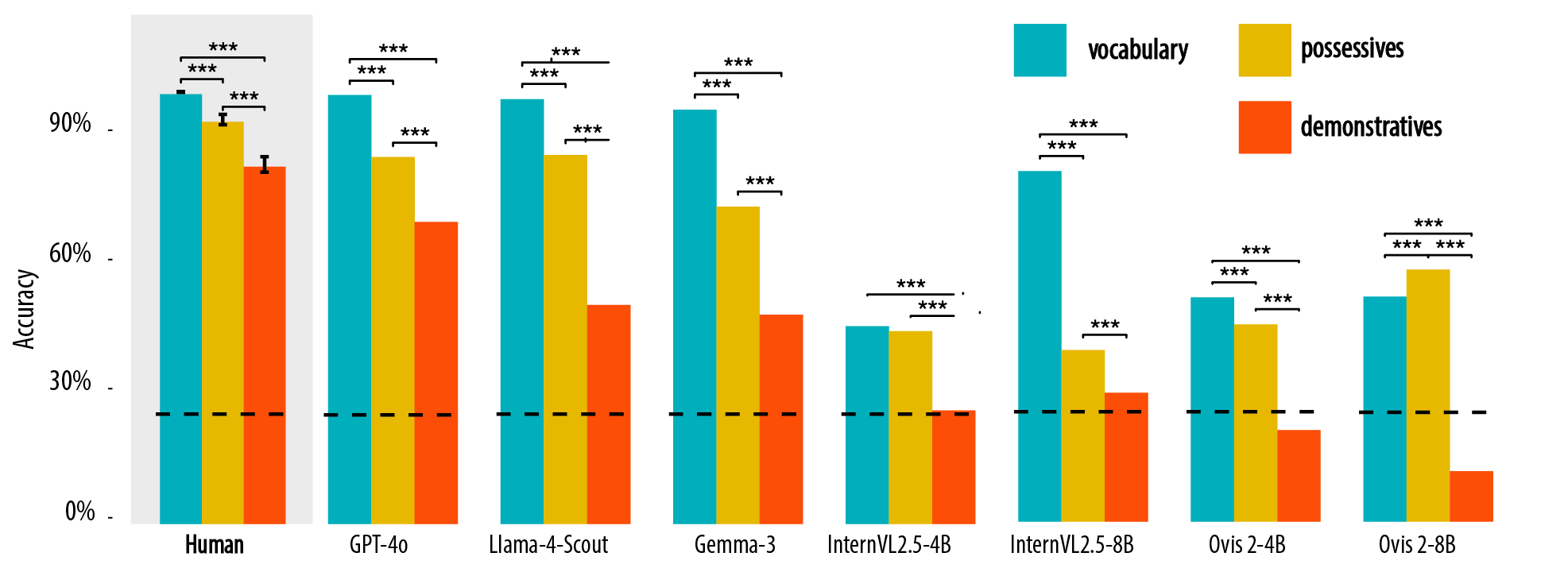}
\caption{\textbf{Human and MLMs production task accuracy}. Error bars represent ±1 standard error across human participants. Since models were evaluated deterministically, no error bars are shown for MLMs. Significance (*) was established via logistic regression with Task as predictor, followed by post-hoc pairwise comparisons using estimated marginal means. The dashed line indicates chance level. 
}
\label{figure-2}
\end{figure*}
\paragraph{In-context learning and Instruction-based prompting} We here evaluated two widely used prompting strategies to enhance MLMs’ performance in tasks involving perspectival words \cite{wei2022chain, qin-eisner-2021-learning}. First, in in-context learning, we presented four annotated exemplars—one for each answer type—prior to the task. Second, in instruction-based prompting, we crafted prompts that emphasized relevant linguistic cues (e.g., speaker identity, preference, ownership, spatial proximity) and encouraged the models to produce step-by-step reasoning along with their responses. All prompts were customized for the specific task. We further examined the error patterns that emerged under these strategies and compared them with those observed in the main experiment. More details are provided in Appendix~\ref{learning}.
\section{Results}
\subsection{Human vs. MLM Performance on Vocabulary and Perspectival Words}
Figure~\ref{figure-2} reports task accuracy for human participants and seven MLMs across the three tasks. We first established the human baseline and then evaluated MLM performance relative to it.

Despite overall high accuracy, human participants showed systematic performance differences across tasks: Vocabulary elicited the highest accuracy, followed by Possessives, with Demonstratives exhibiting the lowest performance. This reduction in task accuracy confirms that vocabulary and perspectival words pose different cognitive demands on language users. Within perspectival words, we interpret the performance difference as an effect of cognitive complexity, reflecting the semantics of these words \citep{levinson2006deixis}: possessive use relies on a binary object categorization (as either self-owned or other-owned), whereas demonstrative use requires distinguishing objects along a continuous spatial dimension (based on their relative proximity to the speaker). Supporting this interpretation, a comprehension version of our task revealed that possessives (`mine' vs. `yours') show binary rating patterns—high for target objects, low for non-targets—while demonstratives (`this' vs. `that') exhibit proximity-based gradients, with `this' showing sharper distance sensitivity than `that' (See Appendix~\ref{comprehension} for full human comprehension experiments and results).

A similar hierarchy of task performance was observed across most models. Five of the seven tested models—GPT-4o, Llama-4-Scout, Gemma-3, InternVL2.5-8B-MPO, and Ovis2-4B—achieve the highest accuracy on Vocabulary, intermediate accuracy on Possessives, and the lowest accuracy on Demonstratives. The remaining two models (InternVL2.5-4B-MPO and Ovis2-8B) show broadly comparable trends, except for deviations in the Vocabulary task due to notably poor performance, primarily stemming from difficulties in using personal pronouns (see Appendix~\ref{error} for a detailed error analysis).


\begin{figure*}[h]
\centering
\includegraphics[width=1\textwidth]{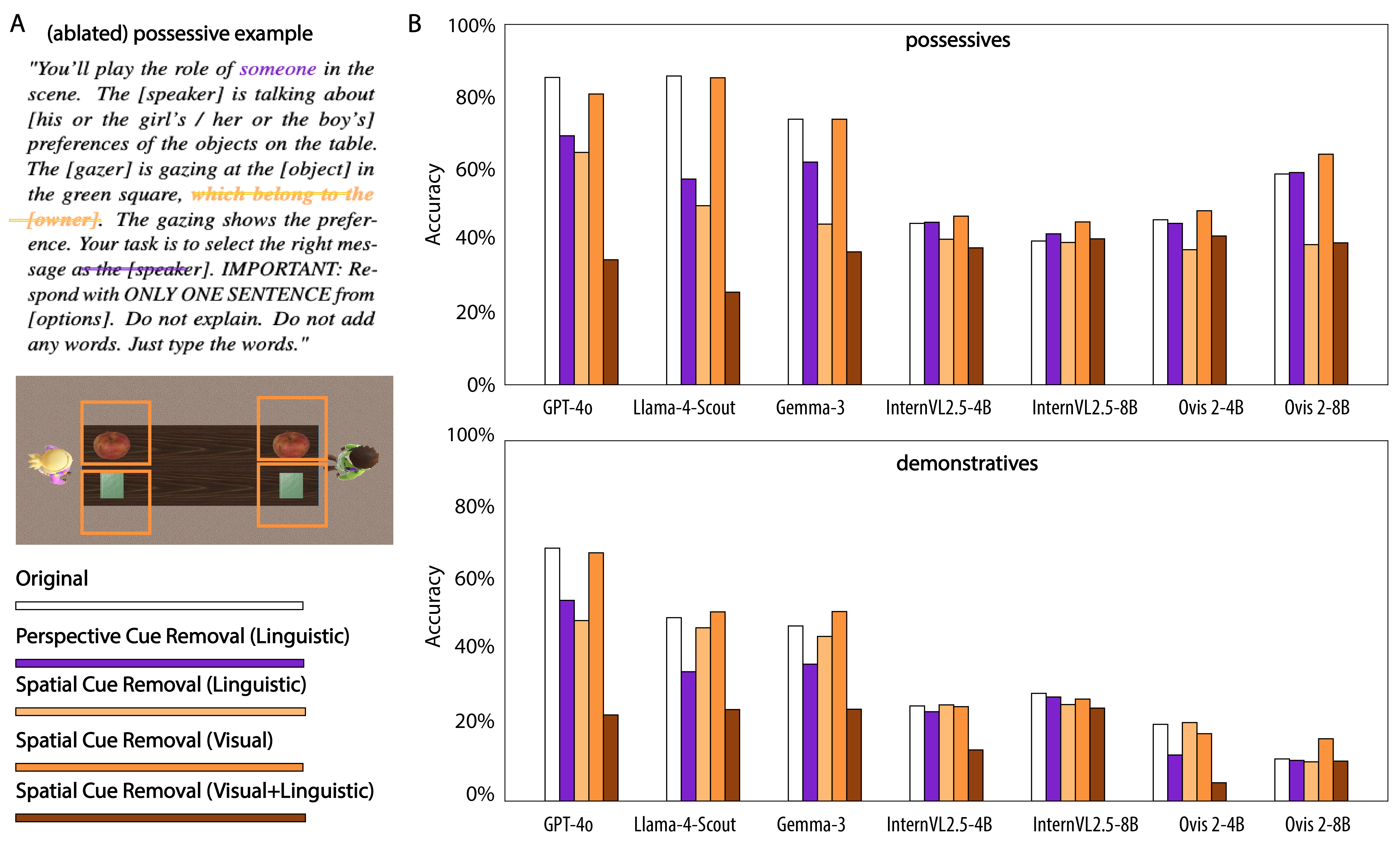}
\caption{\textbf{Ablation analysis of perspectival and spatial cues in possessive and demonstrative reference production across MLMs.} Performance is shown for the original task and four ablation conditions: \textit{Perspectival Cue Removal (Linguistic)}, where model viewpoint (boy vs. girl) is removed from the prompt; \textit{Spatial Cue Removal (Linguistic)}, where ownership markers (possessives) or proximity terms (demonstratives) are omitted; \textit{Spatial Cue Removal (Visual)}, where all objects appear inside identical green squares; and \textit{Spatial Cue Removal (Visual+Linguistic)}, combining both spatial manipulations. \textbf{(A)} Example of an ablated possessive trial (see Appendix~\ref{app:ablation} for complete ablation materials). \textbf{(B)} Top panel: Possessive ablated task accuracy; Bottom panel: Demonstrative ablated task accuracy.}
\label{figure-3}
\end{figure*}

Next, we compared human and MLM performance across tasks. On the Vocabulary task, most models approach ceiling performance comparable to humans, indicating that vocabulary word selection poses minimal difficulty in current MLMs. In contrast, a substantial performance gap emerges in the Possessive and the Demonstrative tasks compared with humans. Even the top-performing commercial models (GPT-4o, Llama-4-Scout, and Gemma-3) achieve accuracy scores that remain well below human performance, despite humans themselves not attaining perfect accuracy on these tasks. For the Possessive task, humans achieve 93\% accuracy compared to the best model (Llama-4-Scout) at 85.7\%, while the gap is more pronounced for the Demonstrative task, with humans at 83\% versus the best model (GPT-4o) at only 70.2\%. The remaining models exhibit even greater deficits, with markedly lower accuracy on the Possessive task and near-chance performance on Demonstratives. These results reveal a clear divergence between MLMs and humans in perspectival word use, contrasting with their comparable performance on vocabulary word selection.

\subsection{Where Models Fall Short: Ablations on Spatial and Perspectival Cues}
The results above show that MLMs struggle with perspectival words more than humans—but where exactly do models fall short? Perspectival words are grammatical forms whose correct use in our task depends on integrating two types of information: perspectival cues (identifying the speaker and listener) and spatial cues (determining the referent’s relative position to the speaker). We hypothesize that successfully combining these cues requires perspective-taking and spatial reasoning—cognitive capacities that likely account for the difficulties MLMs encounter relative to humans.

To identify the source of these deficits, we conducted an ablation study examining how models use and integrate perspective and spatial cues when producing possessive and demonstrative pronouns. We systematically manipulated the availability of perspective and spatial information to determine the origin of the observed difficulties (See an ablated example in Figure~\ref{figure-3}\textbf{A}).

We first examined the effect of removing perspectival cues for both tasks (Figure~\ref{figure-3}\textbf{B}), comparing the original condition with the \textit{Perspectival Cue Removal (Linguistic)} condition. Models show a clear split in their ability to use perspective information. The top-performing models (GPT-4o, Llama-4-Scout, Gemma-3) demonstrate clear accuracy drops when perspectival cues (indicating which role they play in the scene) were removed, suggesting their reliance on viewpoint information to select the correct perspectival term. In contrast, the remaining models showed minimal change—or even slight improvements—after the removal of perspectival cues, indicating that they either fail to leverage these cues or are misled by them. These results reveal a divergence in perspective-taking capabilities: advanced models demonstrate sensitivity to shifts in speaker–listener perspectives comparable to humans, whereas other models appear insensitive to—or unable to properly integrate—perspectival information. This difference may relate to training data, as commercial models are likely exposed to more conversational contexts requiring perspective tracking across turns—a question that deserves further investigation.

We next evaluated how models use and integrate spatial cues from linguistic and visual inputs (Figure~\ref{figure-3}, comparing the original condition with three ablation conditions: \textit{Spatial Cue Removal (Linguistic)}, \textit{Spatial Cue Removal (Visual)}, and \textit{Spatial Cue Removal (Visual+Linguistic)}). Removing spatial cues from the linguistic prompt (\textit{Spatial Cue Removal (Linguistic)}) decreases task performance overall but produces markedly different effects across the two tasks. For Possessives, all models exhibit substantial accuracy drops, indicating their critical reliance on ownership markers in the prompt. For Demonstratives, however, only GPT-4o shows a clear decline. This trend aligns with the differing spatial semantics of these word classes: possessives rely on binary categorization (self-owned vs. other-owned) that can be explicitly specified linguistically, whereas demonstratives encode proximity-based gradients (near vs. far) that are more difficult to capture through linguistic description alone.

In contrast, removing visual spatial cues alone (\textit{Spatial Cue Removal (Visual)}) has minimal impact on task accuracy: all models maintain—or even slightly improve—their performance, indicating that visual information provides little benefit on its own and can sometimes mislead when processed without linguistic support. Taken together, these results suggest that MLMs rely more heavily on linguistic than visual input, a trend consistent with previous findings on modality imbalance in these models \citep{deng2025words,wu2025language}.

In the combined removal condition (\textit{Spatial Cue Removal (Visual+Linguistic)}), performance drops substantially more for both possessive and demonstrative tasks than when only linguistic cues are removed, demonstrating that visual information contributes to model performance—but only when paired with linguistic context. This reveals a critical bias: MLMs rely on linguistic input as the primary scaffold for spatial reasoning, with visual information refining (rather than independently establishing) spatial relationships. In contrast, humans integrate visual and linguistic spatial information in parallel during conversation \citep{kita2003does,rubio2022demonstrative}, highlighting a fundamental difference in multimodal processing that warrants further investigation in future model development.


\subsection{Bridging the Gap: Toward More Human-like Referential Expressions}

\begin{figure*}[ht]
\centering
\includegraphics[width=1\textwidth]{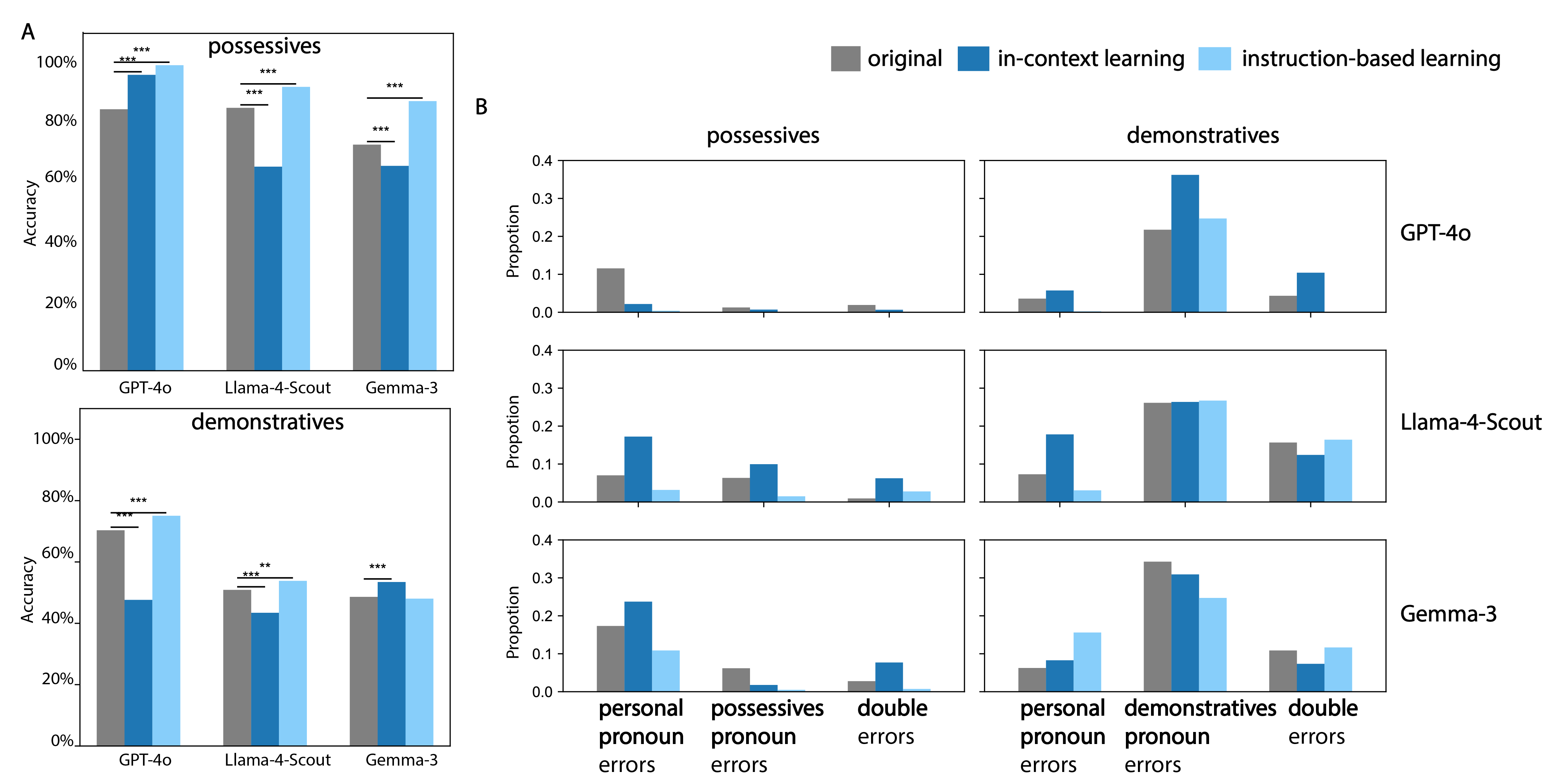}
\caption{\textbf{Effect of prompt approaches on MLM performance and error patterns.} \textbf{A:} Task accuracy for top-performing models (GPT-4o, Llama-4-Scout, Gemma-3) across three prompt conditions: original zero-shot, in-context learning with examples, and instruction-based prompting. Significance (*) was established via logistic regression analysis with response correctness as outcome and Prompt Approach as predictor, followed by estimated marginal means pairwise comparisons. \textbf{B:} Error type distributions across three conditions}
\label{figure-4}
\end{figure*}




In the previous section, we showed a clear performance gap between humans and models in perspectival word use, even for the advanced MLMs (GPT-4o, Llama-4-Scout, and Gemma-3). To improve model alignment with human-level competence on the Possessive and Demonstrative tasks, we explored two widely used approaches with these models: (1) instruction-based prompting, where models reason through explicit pragmatic rules, and (2) in-context learning, where models learn from solved examples (see implementation details in Appendix~\ref{learning}). Figure~\ref{figure-4}\textbf{A} presents accuracy and error patterns for possessives and demonstratives, comparing original performance with in-context and instruction-based prompting conditions.

For the Possesive task, instruction-based prompting substantially enhance performance for all three models, with explicit prompts directing models to consider who is speaking, who expresses a preference, and who owns the object. This approach achieves improvements over the original condition and approaches human-level accuracy. In contrast, in-context learning—providing solved examples without explicit reasoning—proves markedly less effective: only GPT-4o shows a significant improvement, while the other two models exhibit notable performance decreases. Error analysis (Figure~\ref{figure-4}\textbf{B}) shows that gains from instruction-based prompting primarily stem from reductions in possessive pronoun errors, reflecting improved understanding of ownership distinctions when explicitly guided.

However, for the Demonstrative task, even with similar prompts plus additional guidance on the referent’s distance to the speaker, models continue to struggle. Only GPT-4o and Llama-4-Scout show marginal improvements, and their overall performance remains well below human levels. In-context learning generally yields worse results, except for Gemma-3, which exhibits a modest improvement over the original condition. Error analysis (Figure \ref{figure-4}) indicates that performance gains mainly reflect improvements in perspective-taking (reductions in personal pronoun errors) rather than enhanced spatial reasoning (reductions in demonstrative pronoun errors). Demonstrative pronoun errors persist, highlighting that while perspective-taking can be improved through explicit instructional prompts, the spatial reasoning required for demonstrative use remains a major challenge for current MLMs.


\section{Discussion}
In this work, we propose a unified framework for directly comparing the usage of three word classes in referential communication. Our results reveal a clear difficulty hierarchy: Vocabulary < Possessives < Demonstratives. Vocabulary words rely on stable object-label associations, possessive pronouns require perspective-taking to anchor reference to the speaker's viewpoint, and demonstratives additionally demand fine-grained spatial reasoning for proximity-based distinctions.

Critically, this hierarchy is far more pronounced for MLMs than humans, revealing fundamental limitations in perspective-taking and spatial reasoning in these models. Instruction-based prompting substantially improves possessive production, bringing model accuracy near human levels through explicit pragmatic scaffolding. However, demonstratives show no comparable gains—even with explicit spatial guidance, performance remains well below the human benchmark. This asymmetry suggests that while perspective-taking can be enhanced through instruction, spatial reasoning deficits represent a persistent challenge for current MLMs.

We hypothesize that successful demonstrative use may require embodied, interactive communication in physical environments \citep{peeters2016and}—experiences through which humans, but not current MLMs, acquire spatial language. Future work should investigate whether models trained with richer spatial interaction data or equipped with explicit spatial reasoning modules can bridge this persistent gap in referential communication. In doing so, future studies could provide insights into MLMs' social micro-processes \citep{rubio2024tracking}.
\section{Limitations}
First, we employed a simple multiple-choice task within a single, highly controlled scenario to facilitate direct comparison between models and humans. While this approach provides clear metrics, it does not capture the full complexity of perspectival word usage in face-to-face communication \citep{holler2025facial}. The fixed scenario isolates differences in the lexical use of three perspectival word types by minimizing irrelevant variability (e.g., variance in model visual perception) and ensuring that performance differences reflect the words themselves rather than scene complexity. Although it captures core ingredients of face-to-face perspectival communication (e.g., speaker, addressee, gaze, near/far relations), its simplicity may limit generalizability to more complex settings; extending to additional scenarios is a key direction for future work. Moreover, the forced-choice paradigm is not intended to assess Theory of Mind, social micro-processes, or full pragmatic competence, but rather to isolate the integration of pragmatic and spatial cues required for producing perspectival expressions. Accordingly, our findings should not be taken as evidence of general deficits in these capacities, but as indicating difficulties with pragmatic and spatial cues integration. Future work should examine how these findings relate to direct evaluations of Theory of Mind, social interaction processes, and pragmatic reasoning. Second, we only tested two strategies for improving models' production of human-like perspectival words. Other methods remain unexplored, such as persona-based prompting \citep{tan2024phantom}, hyperparameter optimization \citep{oliver2024crafting}, and advanced reasoning frameworks \citep{wei2022chain,yao2023tree}, which future work can further investigate. Third, due to computational resource limitations, we tested only four small models (Interns and Ovis <10B) in addition to models accessed via APIs. Finally, we evaluated models exclusively on English language materials, with our task design being based on English perspectival word usage. We acknowledge that perspectival word usage varies across languages and cultures \citep{rubio2022demonstrative,jara2024demonstratives,levinson2003space,levinson2018introduction,majid2004can,tenbrink2004identifying}. Future studies should examine these phenomena across diverse linguistic and cultural contexts. 
\section{Ethical considerations}
Our research followed ethical guidelines throughout. We recruited monolingual native English speakers through Prolific, obtained approval from our institutional review board, and kept all participant data anonymous and secure. Understanding what multimodal language models can and cannot do matters for researchers and the public alike. Our work offers new ways to test how different these models are from humans in handling perspectival words that require pragmatics and social cognition. We hope this approach will increase researchers' interest in evaluating human-centric capabilities of these models and leads to improvements that make them more useful for everyday conversations.

\section{Acknowledgments}
This study was supported by the Multimodal Language Department at the Max Planck Institute for Psycholinguistics in Nijmegen. DD is funded by an IMPRS fellowship. AW is supported by the NSTC. We thank Esam Ghaleb for his feedback.
\bibliography{anthology}
\bibliographystyle{acl_natbib}

\appendix
\clearpage
\newpage
\section*{Appendix}
\label{sec:appendix}
\appendix

\section{Sanity-check: MLMs' Performance on Object Recognition and Speaker/Gazer Identification}
\label{app:sanity}

In the human experiments, the speaker was indicated with a speech bubble and their preference with an intermittent gaze line connecting the eyes to the target object. In the MLM experiments, these cues were instead conveyed in text prompts, with the only visual cue being a green box around the target object.

We did not use the human-adapted materials for MLMs because models may not interpret speech bubbles or gaze lines as humans do (See more details in the \textbf{Human Experiment Setting Check} below). To avoid confounds, we provided information primarily in text, supplemented with a simple visual cue. Conversely, humans did not receive written descriptions of speaker or gazer identity, as this would burden working memory compared to perceiving the cues directly in the display (e.g., in some trials, participants would have to remember that they boy is the speaker and the girl is the gazer while looking at a bare display, rather than receiving that information from the speech bubble and line of gaze in the actual display).

In short, we treated visual information as the native modality for humans and written text as the native modality for models. To verify comparability, we first ran a sanity check by testing MLMs on the original human stimuli.

\paragraph{Human Experiment Setting Check} We tested MLMs on the human experiment stimuli by probing each image three times: object in the green square, speaker, and gazer. As shown in Table~\ref{tab:sanity_check_human}, MLMs struggled—especially with object recognition, indicating that the human-optimized stimuli were not well-suited to the models’ capacities.

\begin{table}[ht]
\centering
\small
\begin{tabular}{@{}llr@{}}
\toprule
\textbf{Test Type} & \textbf{Model} & \textbf{Accuracy} \\
\midrule
\multirow{6}{*}{\parbox{3cm}{Object\\Recognition}} 
& gpt-4o & 0.375 \\
& Llama-4-Scout & 0.438 \\
& Gemma-3 & 0.375 \\
& InternVL2\_5-4B-MPO & 0.438 \\
& InternVL2\_5-8B-MPO & 0.375 \\
& Ovis2-8B & 0.562 \\
\midrule
\multirow{7}{*}{\parbox{3cm}{Gazer\\Identification}} 
& gpt-4o & 0.875 \\
& Llama-4-Scout & 0.938 \\
& Gemma-3 & 0.938 \\
& InternVL2\_5-4B-MPO & 0.750 \\
& InternVL2\_5-8B-MPO & 0.750 \\
& Ovis2-4B & 0.562 \\
& Ovis2-8B & 0.750 \\
\midrule
\multirow{7}{*}{\parbox{3cm}{Speaker\\Identification}} 
& gpt-4o & 1.000 \\
& Llama-4-Scout & 0.938 \\
& Gemma-3 & 0.812 \\
& InternVL2\_5-4B-MPO & 0.750 \\
& InternVL2\_5-8B-MPO & 0.750 \\
& Ovis2-4B & 0.562 \\
& Ovis2-8B & 0.875 \\
\bottomrule
\end{tabular}
\caption{Overall accuracies of MLMs in the Human Experiment Setting.} 
\label{tab:sanity_check_human}
\end{table}

\paragraph{MLM Experiment Setting Check} We next tested MLMs on a subsample of the MLM stimuli, ensuring that each of the 128 objects was queried once. Using the same three recognition tasks (object, speaker, gazer), we obtained the results in Table~\ref{tab:sanity_check_mlm}. MLMs robustly recognized all scene elements, confirming that the adapted stimuli support parallel task designs differing only in modality (visual for humans, text+visual for models). This ensures that divergences in behavior reflect word use rather than confounds in scene understanding.

\begin{table}[ht]
\centering
\small
\begin{tabular}{@{}llr@{}}
\toprule
\textbf{Test Type} & \textbf{Model} & \textbf{Accuracy} \\
\midrule
\multirow{7}{*}{\parbox{3cm}{Object\\Recognition}} 
& gpt-4o & 1.000 \\
& Llama-4-Scout & 0.984 \\
& Gemma-3 & 0.977 \\
& InternVL2\_5-4B-MPO & 0.969 \\
& InternVL2\_5-8B-MPO & 0.961 \\
& Ovis2-4B & 1.000 \\
& Ovis2-8B & 0.992 \\
\midrule
\multirow{7}{*}{\parbox{3cm}{Gazer\\Identification}} 
& gpt-4o & 0.906 \\
& Llama-4-Scout & 0.992 \\
& Gemma-3 & 1.000 \\
& InternVL2\_5-4B-MPO & 0.992 \\
& InternVL2\_5-8B-MPO & 0.953 \\
& Ovis2-4B & 1.000 \\
& Ovis2-8B & 0.961 \\
\midrule
\multirow{7}{*}{\parbox{3cm}{Speaker\\Identification}} 
& gpt-4o & 1.000 \\
& Llama-4-Scout & 1.000 \\
& Gemma-3 & 1.000 \\
& InternVL2\_5-4B-MPO & 0.984 \\
& InternVL2\_5-8B-MPO & 1.000 \\
& Ovis2-4B & 1.000 \\
& Ovis2-8B & 1.000 \\
\bottomrule
\end{tabular}
\caption{Overall accuracies of MLMs in the MLM Experiment Setting.}
\label{tab:sanity_check_mlm}
\end{table}

\clearpage
\newpage
\section{Tasks for Humans}
\label{app:humans}
Informed consent was obtained from all participants prior to the start of the task. All tasks had an average duration of 6 minutes for more accurate across-task comparisons, and to avoid participant fatigue. Participants were paid 9 GBP per hour, which was evaluated as a ``Good rate'' by Prolific.

In the following sections we provide further details of the tasks described in the main text.

\subsection{Production Task} Participants were instructed to play the role of the speaker and complete the speech bubble by selecting the correct message (from among 4 options) given the object preference shown by one of the characters through their line of gaze. Participants were shown two sample trials in each task.

\begin{figure}[t]
\begin{center}
\includegraphics[width=0.7\columnwidth]{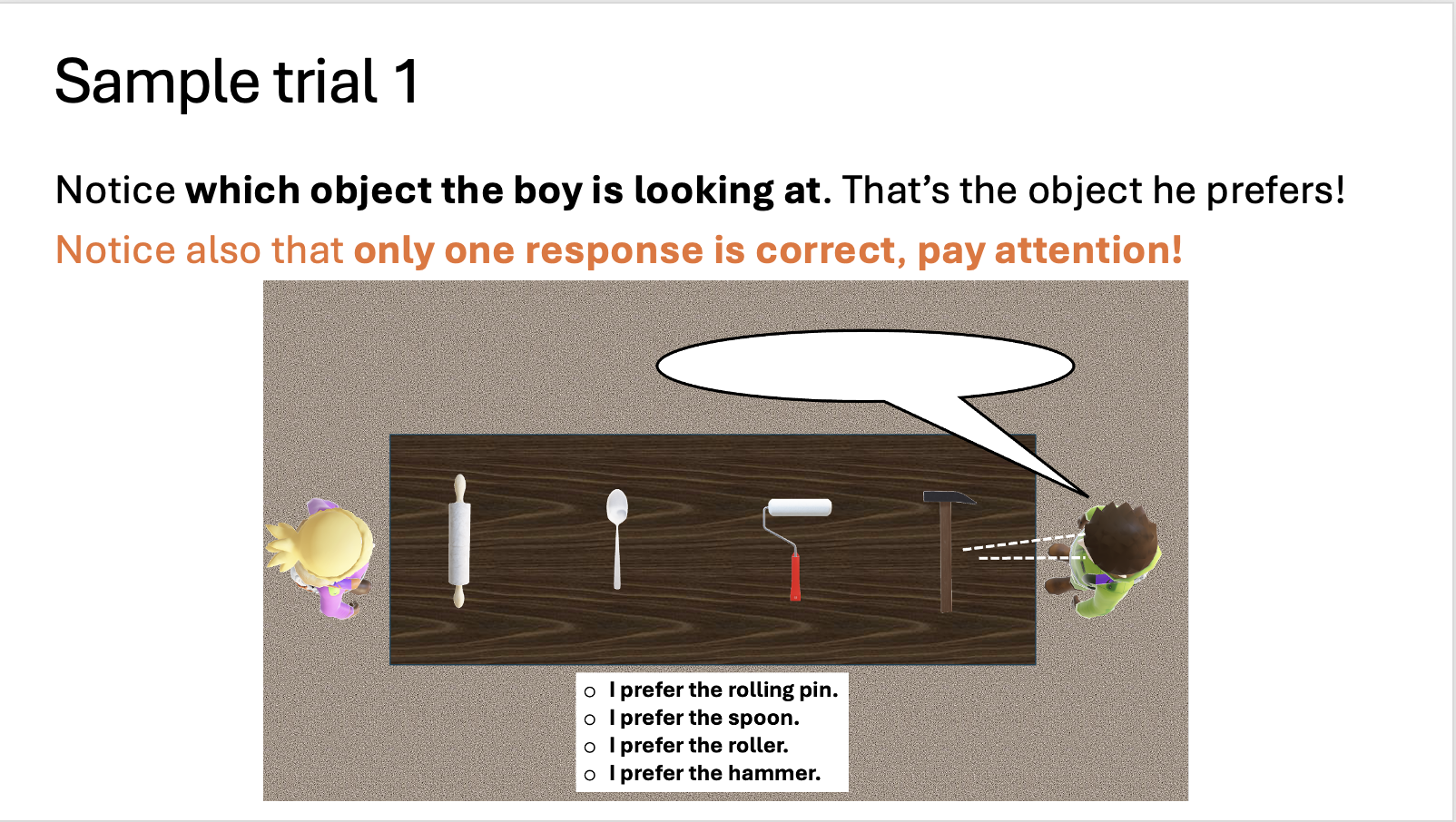}
\end{center}
\label{figure-sample1VOC}
\end{figure}

\begin{figure}[t]
\begin{center}
\includegraphics[width=0.7\columnwidth]{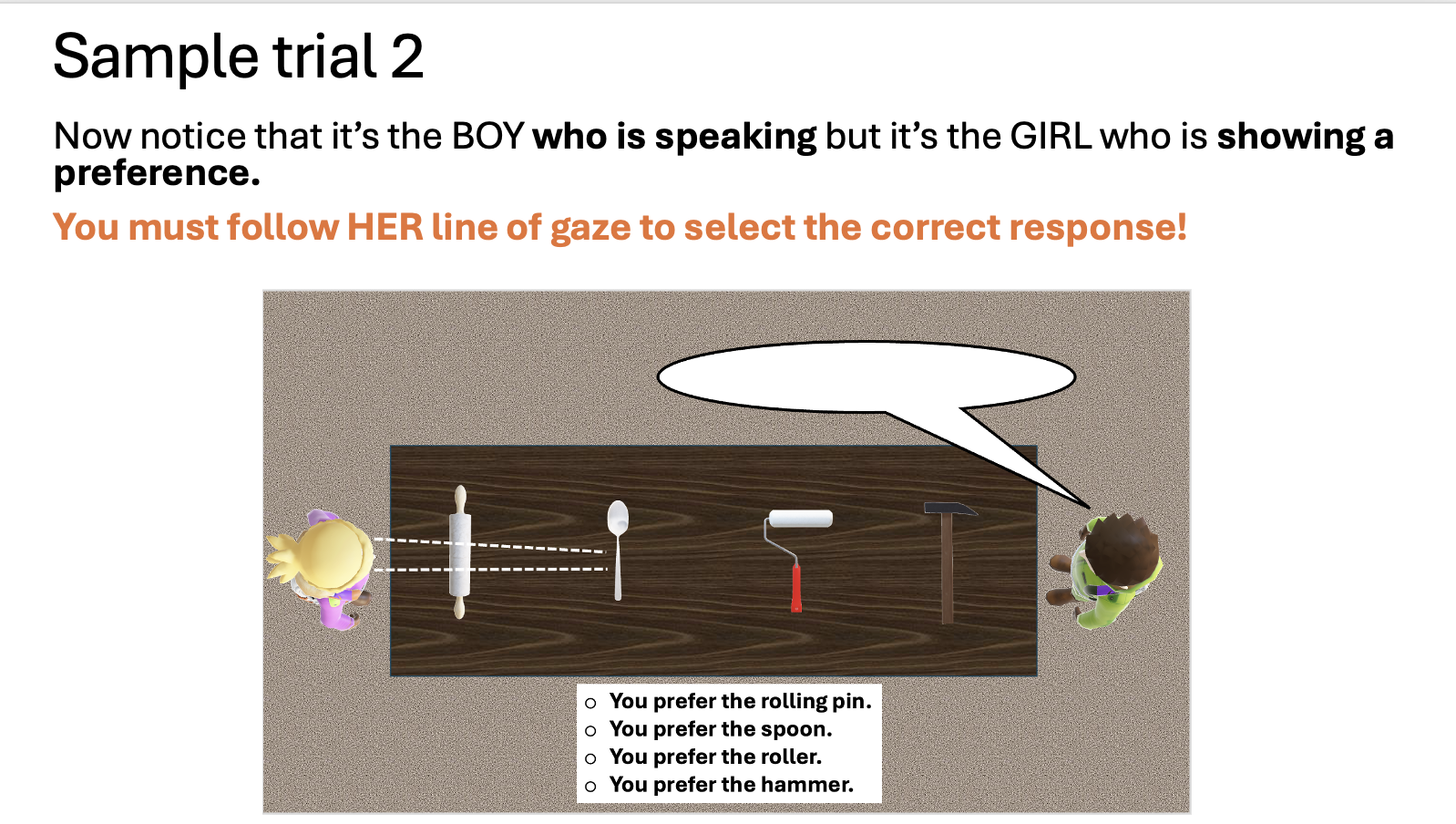}
\end{center}
\caption{Sample trials in the Vocabulary task for humans.}
\label{figure-sample2VOC}
\end{figure}

\begin{figure}[t]
\begin{center}
\includegraphics[width=0.7\columnwidth]{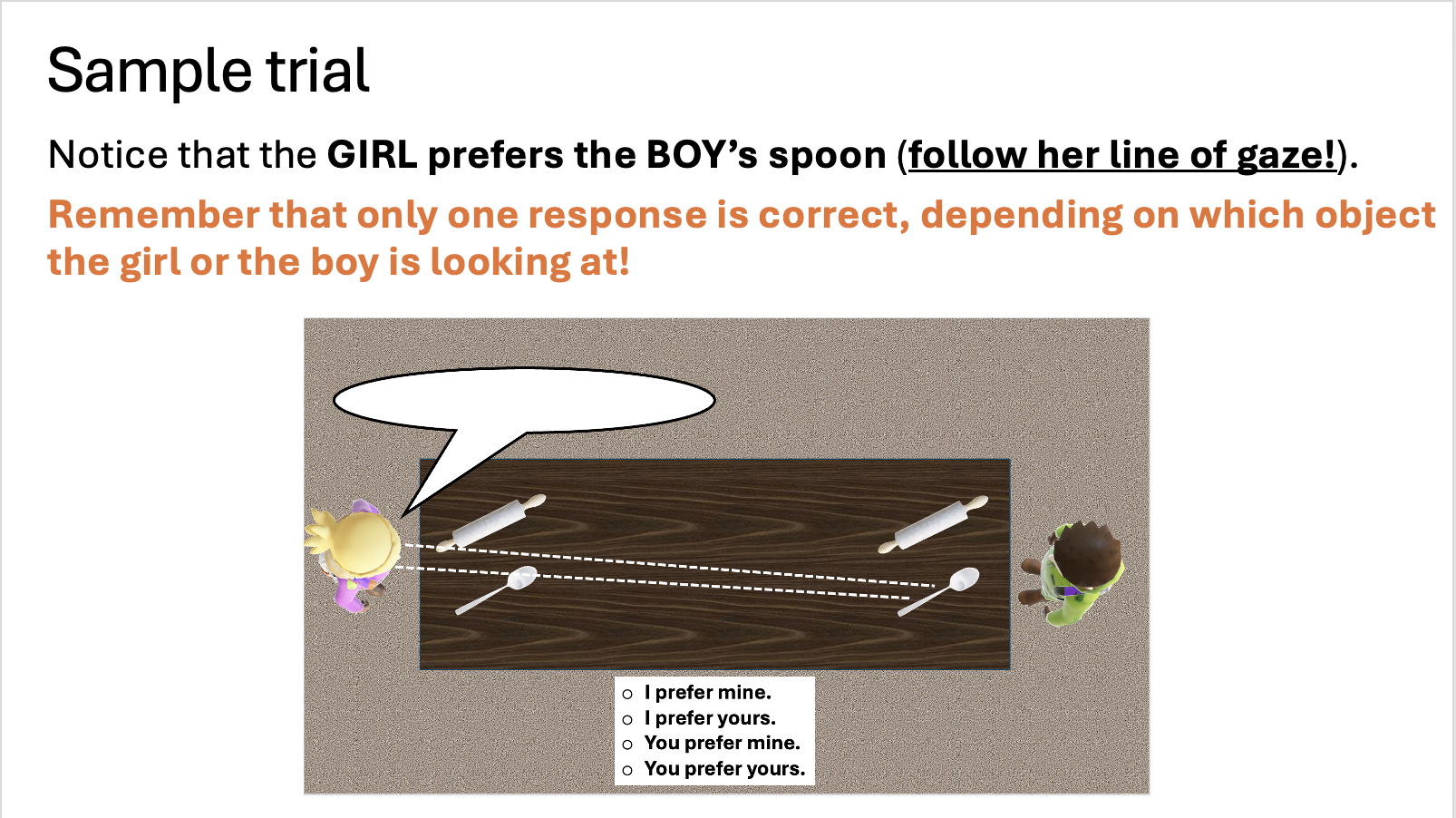}
\end{center}
\label{figure-sample1VOC}
\end{figure}

\begin{figure}[t]
\begin{center}
\includegraphics[width=0.7\columnwidth]{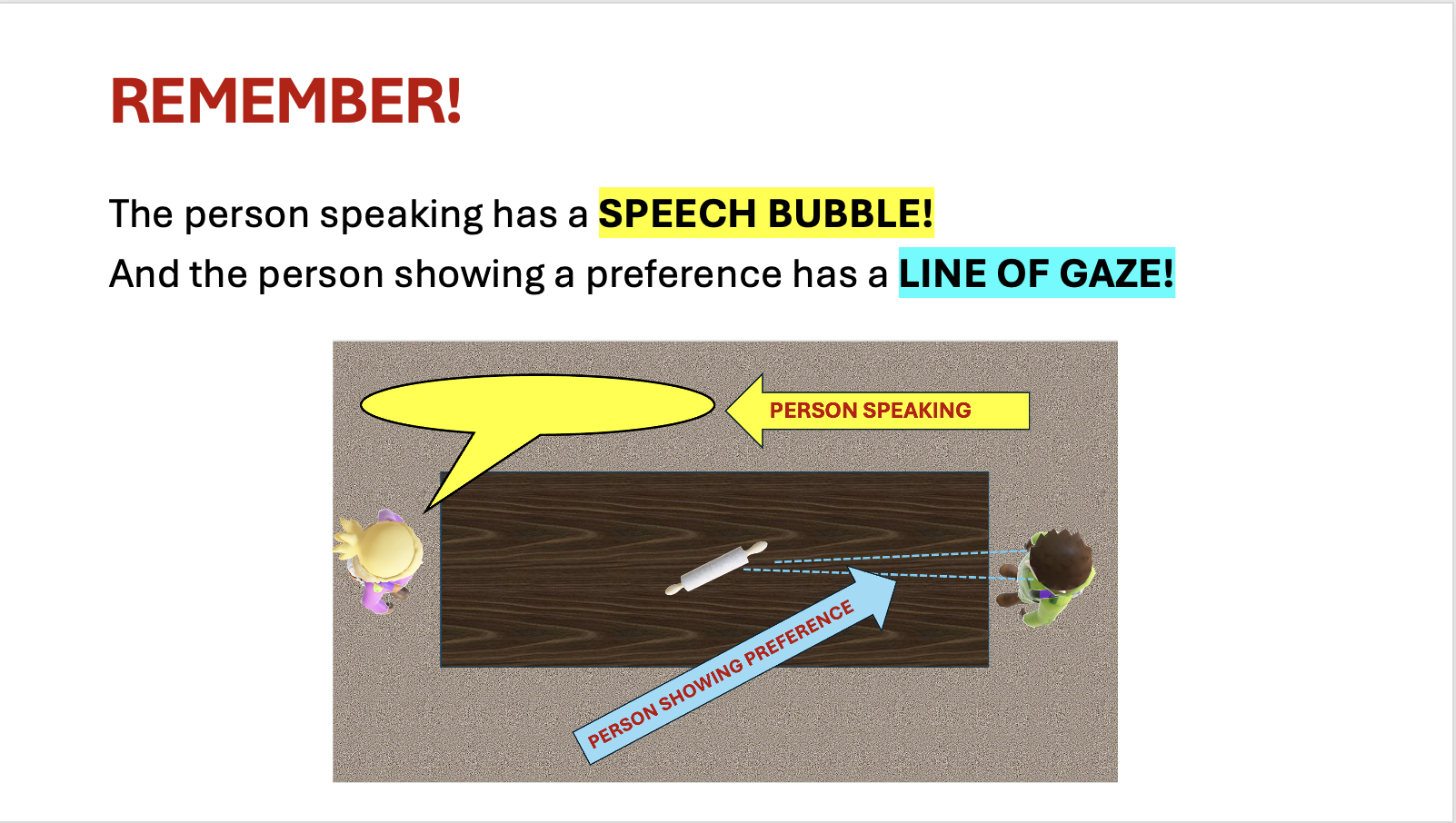}
\end{center}
\caption{Sample trials in the Possessive task for humans.}
\label{figure-sample2VOC}
\end{figure}

\begin{figure}[t]
\begin{center}
\includegraphics[width=0.7\columnwidth]{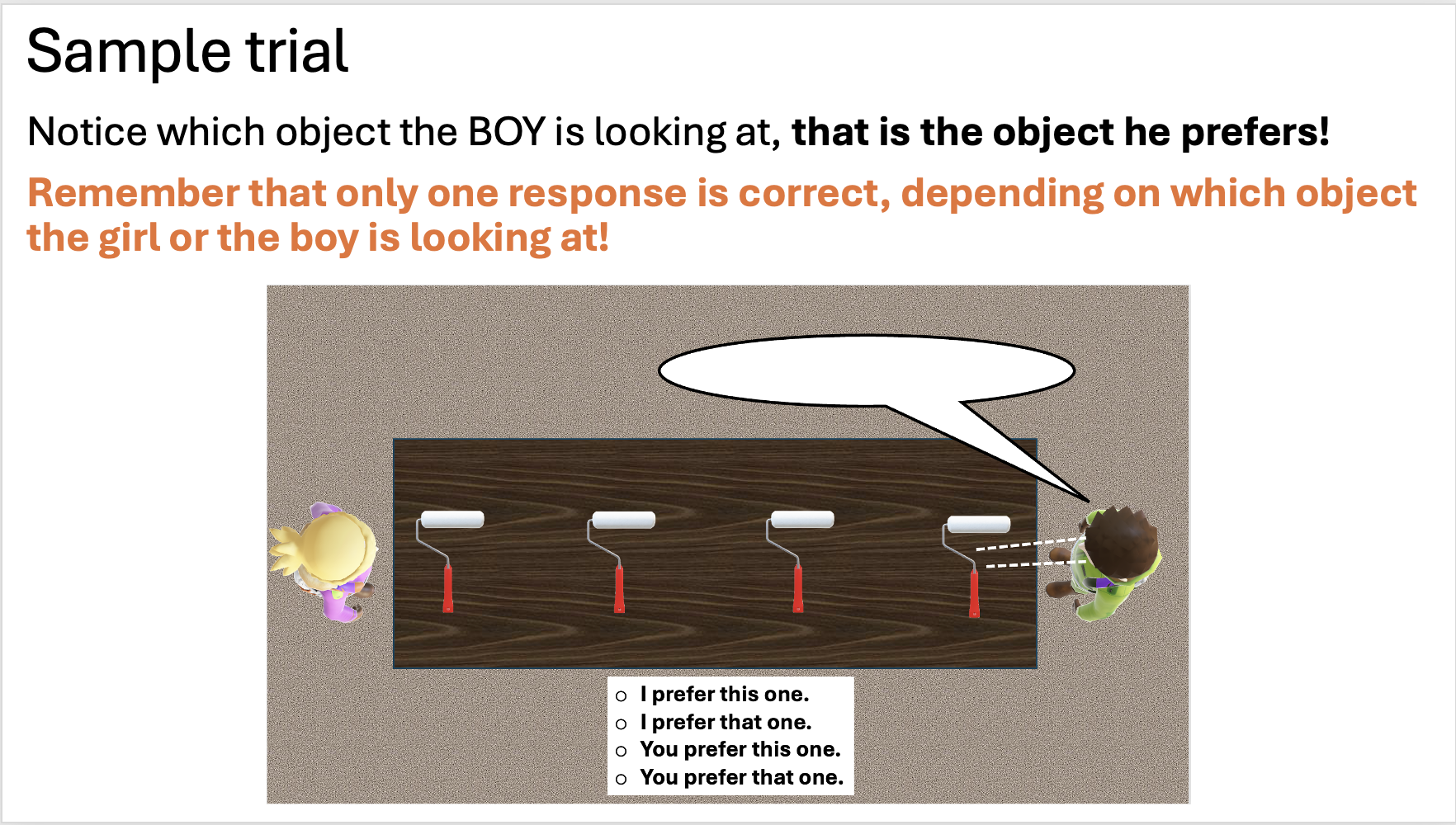}
\end{center}
\label{figure-sample1VOC}
\end{figure}

\begin{figure}[t]
\begin{center}
\includegraphics[width=0.7\columnwidth]{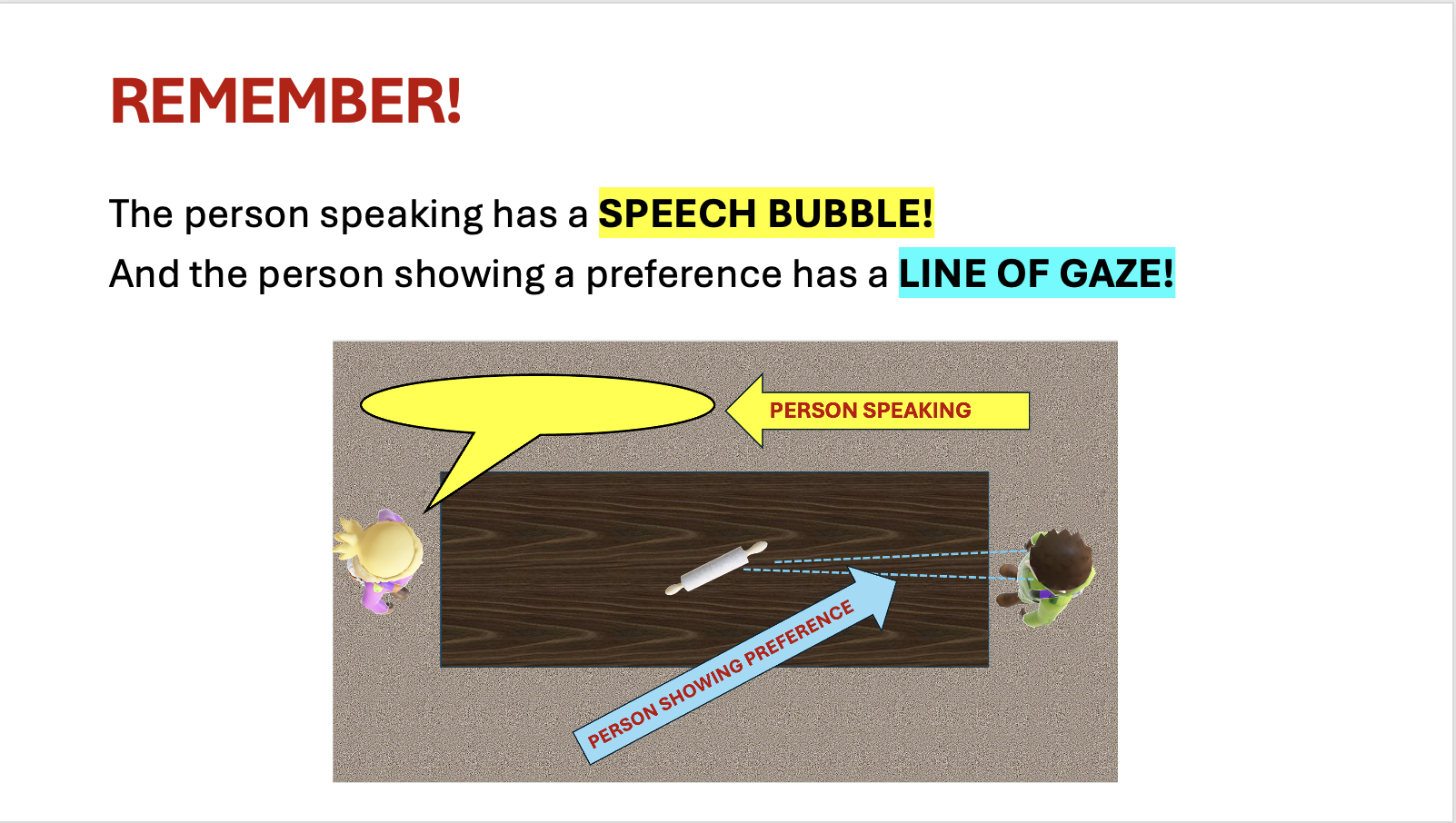}
\end{center}
\caption{Sample trials in the Demonstrative task for humans.}
\label{figure-sample2VOC}
\end{figure}

\subsection{Follow-up Production Task} We aimed to replicate the results of the Production tasks using the same instructions and materials in a within-participants design.

\subsection{Production and Follow-up Production Task Results}

The production task and follow-up production task accuracy are shown in Table \ref{tab:ab_comparison}. We excluded one participant from the production vocabulary task and one from the follow-up production vocabulary task due to random performance. No significant difference in accuracy was detected across the first and follow-up runs of the three tasks.

\begin{table}[t]
  \centering
  \small
  \begin{tabular}{lccc}
    \toprule
    \textbf{Task Type} & \textbf{Production} & \textbf{Follow-up} & \textbf{\textit{p}-value} \\
    \midrule
    Vocabulary                        & 99.84\% & 99.43\%   & 0.0764 \\
    Possessive                      & 93.44\% & 92.44\%   & 0.2037 \\
    Demonstrative                      & 83.00\% & 82.19\%   & 0.4908 \\
    \bottomrule
  \end{tabular}
  \caption{Comparison of production and follow-up production task accuracy.}
  \label{tab:ab_comparison}
  
  \vspace{2pt}
  \raggedright
\end{table}

\clearpage
\newpage
\section{Model Specifications}
\label{app:specs}
We conducted all experiments using GPT-4o, Llama-4-Scout, and Gemma-3 through their respective APIs, locally hosted the remaining models on a standard GPU setup, and completed the full evaluation suite in approximately 40 hours.

\subsection{GPT-4o}
OpenAI’s omni transformer jointly encodes text, vision, and audio in a single network \citep{openai2024gpt4o}.\
The public API exposes a 128\,k token context window.

\subsection{Llama-4-Scout 17B (16E)}
A 17B active-param mixture-of-experts (MoE) auto-regressive LM with early-fusion
MetaCLIP vision branch \citep{meta2025llama4,xu2024demystifying}.\
It is advertised as a 10M-token window.

\subsection{Gemma 3 (27B)}
Decoder-only LM distilled from a larger teacher model and paired with a frozen
400 M-parameter SigLIP encoder at $896{\times}896$ px, adopting a Pan and Scan (P\&S) method for flexible resolutions \citep{gemma3report2025}, providing a 128k token window.

\subsection{InternVL 2.5-4B / 8B (MPO)}
 
InternViT-300M-448px encoder is coupled with different pretrained LLMs (Qwen2.5-3B-Instruct for InternVL 2.5 4B and internlm2\_5-7b-chat for InternVL 2.5 8B) via an MLP projector \citep{chen2024expanding}.\
Adopting Mixed Preference Optimization (MPO) on the Multi-Modal Preference (MMPR) dataset, InternVL 2.5 is the first open model to pass 70 \% MMMU. 

\subsection{Ovis 2-4B / 8B}
Ovis 2-4B and 8B are separately based on Qwen2.5-3B-Instruct/ Qwen2.5-7B-Instruct LLM, with implementing structural embedding alignment by inserting a learnable visual-embedding
table inside an aimv2-huge-patch14-448 vision encoder, mirroring word-token structure {\citep{lu2024ovis}.\
All sizes support videos and multi-image processing.

\clearpage
\newpage
\section{Trial Augmentation and Matching with Human Evaluation}
\label{app:augmentation}
In each model experiment, we augmented the trials to 6144 while retaining alignment to the human evaluations by varying task-irrelevant factors. To compare with human experiment results, we obtained the MLM experiment accuracy as the main metric. The response time of MLMs can be influenced by network conditions or API provider, therefore it is not selected as the primary measurement of MLM performance.

\paragraph{Vocabulary Task} In both human and model evaluations, we grouped the 128 test objects into 32 groups. In the model evaluation, we showed models each object group in half of the possible orders on the table, randomly sampled, which rendered the trials = 6144 = 32 (groups) $\times$ 4 (possible targets) $\times$ 12 (possible orders) $\times$ 2 (possible speakers) $\times$ 2 (possible perspectives). In the human evaluation, the trial number is 16 = 4 (groups) $\times$ 1 (possible targets) $\times$ 2 (possible speakers) $\times$ 2 (possible perspectives).

\paragraph{Possessive Task} In model evaluation, for each object group, we had 6 possible two-object combinations, and we placed the objects in two spatial arrangements, rendering the trials = 6144 = 32 (groups) $\times$ 4 (possible targets) $\times$ 6 (two-object combinations) $\times$ 2 (spatial arrangements) $\times$ 2 (possible speakers) $\times$ 2 (possible gazers). In the human evaluation, the trial number is 32 = 4 (groups) $\times$ 2 (two-object combinations) $\times$ 1 (possible targets) $\times$ 2 (possible speakers) $\times$ 2 (possible perspectives).

\paragraph{Demonstrative Task} In the model demonstrative task, we made 6 scenes in which the two characters stood on either side of the table. The model task trial number is 6144 = 128 (objects) $\times$ 2 (possible targets) $\times$ 6 (scenes) $\times$ 2 (possible speakers) $\times$ 2 (possible gazers). In the human evaluation, the trial number is 32 = 8 (objects) $\times$ 2 (possible speakers) $\times$ 2 (possible perspectives).

\paragraph{Object-Group Effects on Experiment Accuracy}
We used chi-squared test (table \ref{tab:chi2_results}) to see if models performance in the experiments varied with the objects appeared in the trial images. We do not find significant effect in Possessive or Demonstrative experiments. However, we observed significant results in Vocabulary experiment, which is likely due to the task demanding the model to recognize the object and include the object name in the final answer.

\begin{table}[ht]
\centering
\small
\begin{tabular}{@{}l l c c@{}}
\toprule
\textbf{Experiment} & \textbf{Model} & \textbf{$\chi^2$ (df)} & \textbf{p-value} \\
\midrule
Vocabulary   & GPT-4o              & 173.116 (31) & $<$0.001 \\
& Llama-4-Scout       & 126.189 (31) & $<$0.001 \\
& Gemma-3             & 239.487 (31) & $<$0.001 \\
& InternVL2\_5-4B-MPO & 93.080 (31)  & $<$0.001 \\
& InternVL2\_5-8B-MPO & 600.099 (31) & $<$0.001 \\
& Ovis2-4B            & 71.249 (31)  & $<$0.001 \\
& Ovis2-8B            & 56.744 (31)  & 0.003 \\
\midrule
Possessive   & GPT-4o              & 26.220 (31)  & 0.711 \\
   & Llama-4-Scout       & 35.969 (31)  & 0.247 \\
   & Gemma-3             & 25.112 (31)  & 0.763 \\
   & InternVL2\_5-4B-MPO & 21.418 (31)  & 0.900 \\
   & InternVL2\_5-8B-MPO & 15.860 (31)  & 0.989 \\
   & Ovis2-4B            & 15.021 (31)  & 0.993 \\
   & Ovis2-8B            & 6.984 (31)   & 1.000 \\
\midrule
Demonstrative & GPT-4o              & 14.442 (31)  & 0.995 \\
 & Llama-4-Scout       & 32.695 (31)  & 0.384 \\
 & Gemma-3             & 27.129 (31)  & 0.666 \\
 & InternVL2\_5-4B-MPO & 20.342 (31)  & 0.928 \\
 & InternVL2\_5-8B-MPO & 8.989 (31)   & 1.000 \\
 & Ovis2-4B            & 40.622 (31)  & 0.116 \\
 & Ovis2-8B            & 36.492 (31)  & 0.229 \\
\bottomrule
\end{tabular}
\caption{Chi-square test results for object-group effects on model accuracy across vocabulary, possessive, and demonstrative experiments.}
\label{tab:chi2_results}
\end{table}

\clearpage
\newpage
\section{External FoR is explicitly controlled}
\label{app:ExternalFoR}
As shown in Table~2, models can reliably recover the correct speaker and gazer. To further rule out residual spatial ambiguity, we conducted an additional control analysis. Results show that models also answer spatial questions unrelated to demonstrative usage—such as \textit{``Is the object close to you as the speaker?''}—with near-perfect accuracy (Table~\ref{tab:spatial_control}). 

Taken together, our experimental design and control analyses indicate that models have no difficulty resolving frame of reference (FoR) information outside of demonstrative usage, making it unlikely that the observed errors stem from stimulus-level FoR underspecification.

\begin{table}[h]
\centering
\begin{tabular}{lc}
\hline
\textbf{Model} & \textbf{Accuracy} \\
\hline
gpt-4o & 0.914 \\
Llama-4-Scout & 0.992 \\
Gemma-3 & 1.000 \\
InternVL2.5-4B-MPO & 1.000 \\
InternVL2.5-8B-MPO & 0.969 \\
Ovis2-4B & 1.000 \\
Ovis2-8B & 1.000 \\
\hline
\end{tabular}
\caption{Accuracy on control spatial questions unrelated to demonstrative usage.}
\label{tab:spatial_control}
\end{table}

\clearpage
\newpage
\section{Images and Prompts for MLM Experiment}
\label{app:prompt}

In the following sections we provide our sample images prompts used in the MLM experiments. The square brackets in the prompts signify dynamic variables that change by trial.

\subsection{Main Experiment}
\label{zero-shot}
\paragraph{Vocabulary Task}
\begin{figure}[h]
\begin{center}
\includegraphics[width=0.7\columnwidth]{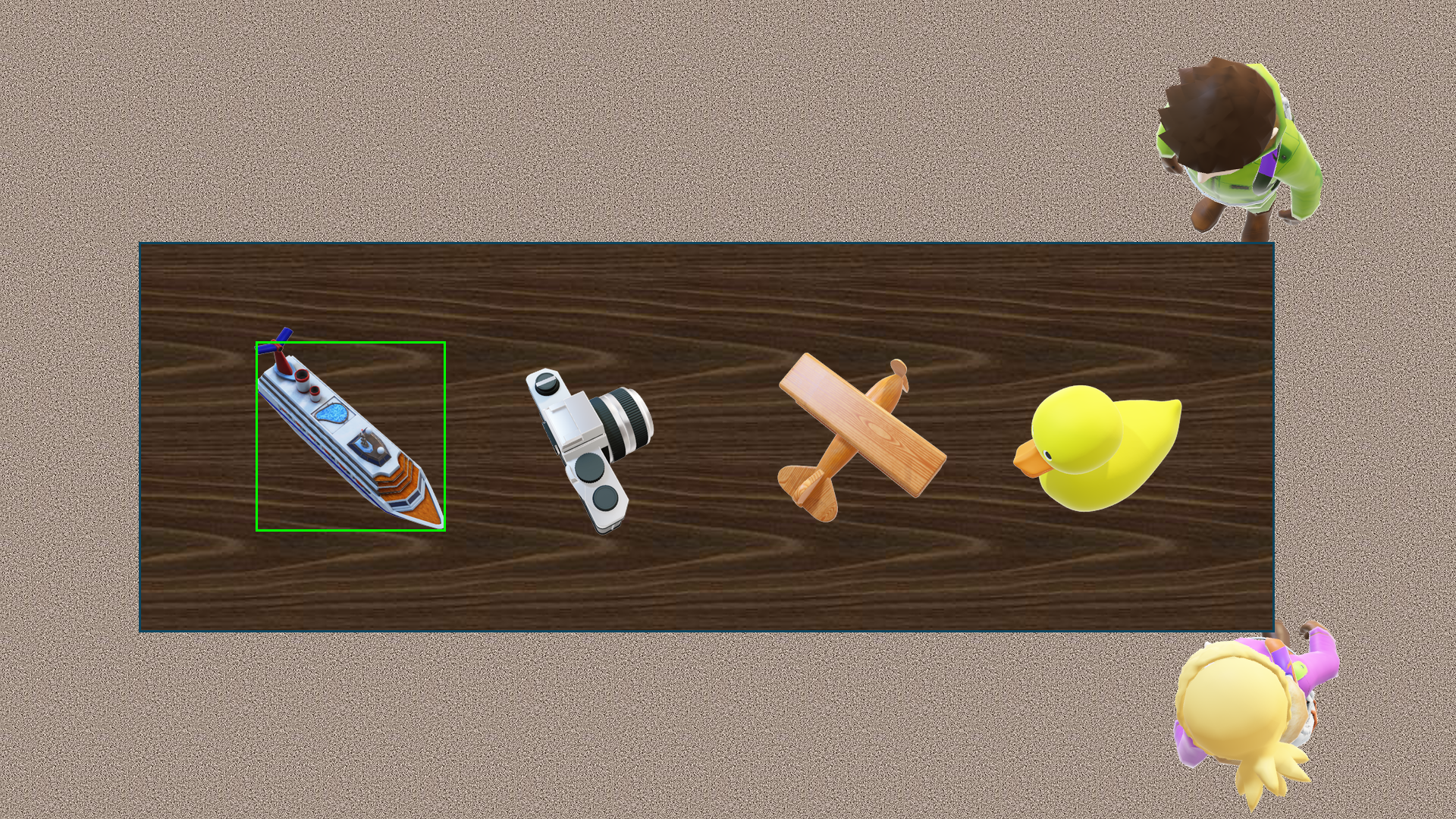}
\end{center}
\caption{Sample image in the Vocabulary task for models.}
\label{figure-sample_model_voc}
\end{figure}
\begin{quote}
    \textit{"You'll play the role of the [speaker] in the scene. The [speaker] is talking about [his or the girl's / her or the boy's] preferences of the objects on the table. The [gazer] is gazing at the [object] in the green square. The gazing shows the preference. Your task is to select the right message as the [speaker]. IMPORTANT: Respond with ONLY ONE SENTENCE from [options]. Do not explain. Do not add any words. Just type the words."}
\end{quote}

\paragraph{Possessive Task}
\begin{figure}[h]
\begin{center}
\includegraphics[width=0.7\columnwidth]{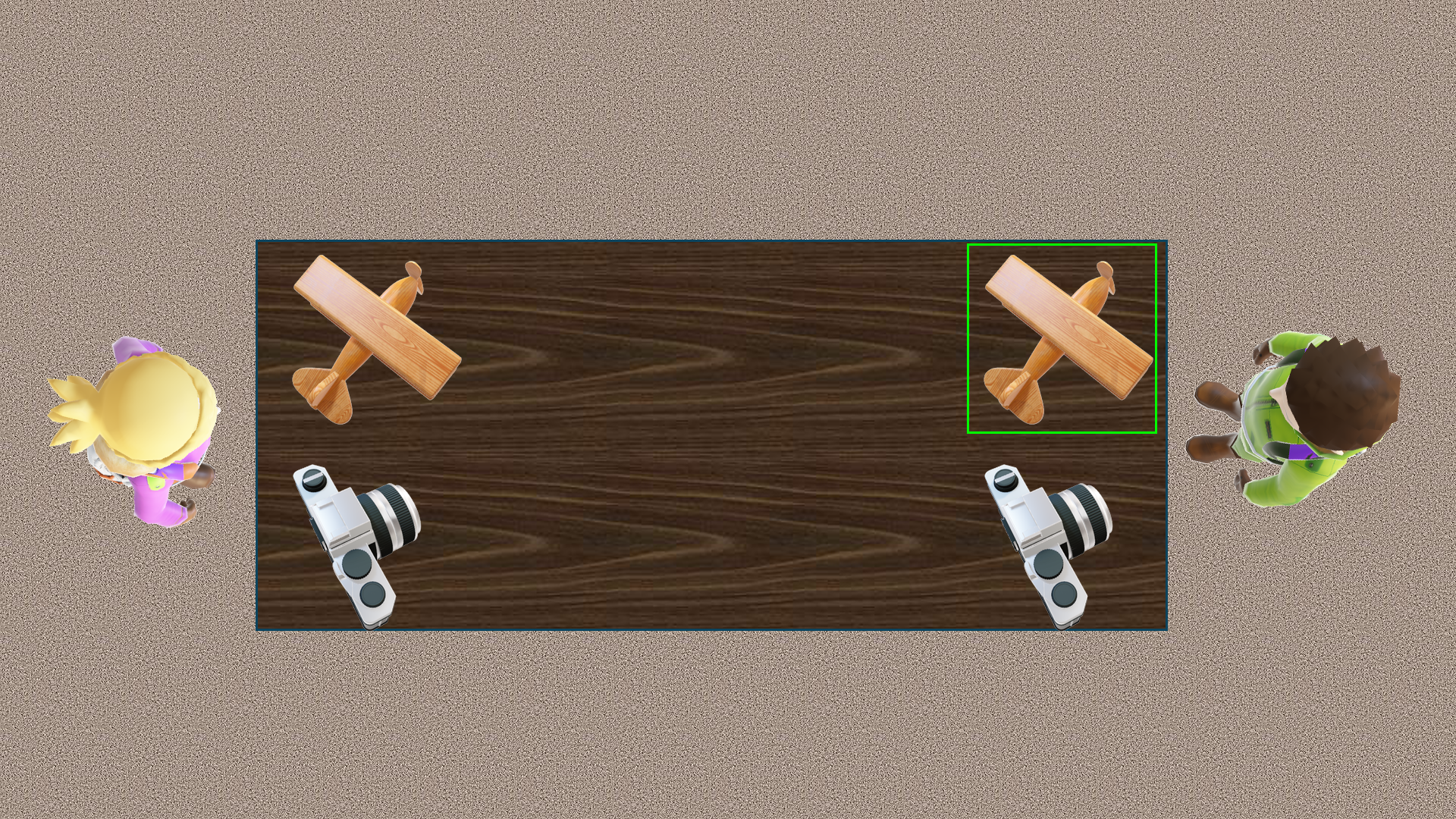}
\end{center}
\caption{Sample image in the Possessive task for models.}
\label{figure-sample_model_pos}
\end{figure}
\begin{quote}
    \textit{"You'll play the role of the [speaker] in the scene. The [speaker] is talking about [his or the girl's / her or the boy's] preferences of the objects on the table. The [gazer] is gazing at the [object] in the green square, \textbf{which belong to the [owner]}. The gazing shows the preference. Your task is to select the right message as the [speaker]. IMPORTANT: Respond with ONLY ONE SENTENCE from [options]. Do not explain. Do not add any words. Just type the words."}
\end{quote}

\paragraph{Demonstrative Task}
\begin{figure}[h]
\begin{center}
\includegraphics[width=0.7\columnwidth]{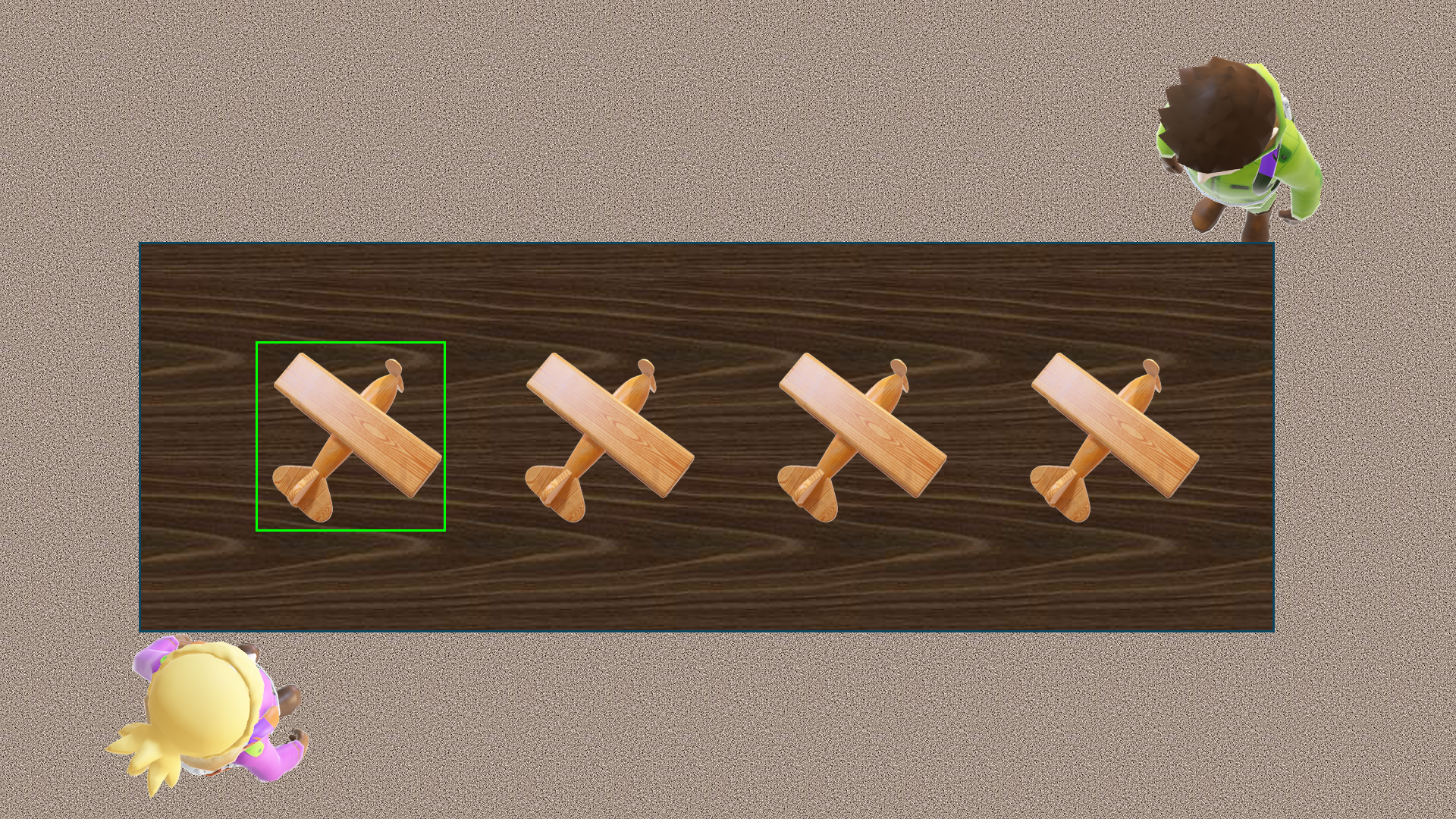}
\end{center}
\caption{Sample image in the Demonstrative task for models.}
\label{figure-sample_model_dem}
\end{figure}
\begin{quote}
    \textit{"You'll play the role of the [speaker] in the scene. The [speaker] is talking about [his or the girl's / her or the boy's] preferences of the objects on the table. The [gazer] is gazing at the [object] in the green square, \textbf{which is close to the [person near target]}. The gazing shows the preference. Your task is to select the right message as the [speaker]. IMPORTANT: Respond with ONLY ONE SENTENCE from [options]. Do not explain. Do not add any words. Just type the words."}
\end{quote}

\subsection{Prompt Variation Experiment}
\label{app:prompt variation}
We provided three prompt variations to assess the robustness of observed effect. The assessment results of the context-first, condensed, and simple prompts are shown in table \ref{tab:prompt_variations_context}, \ref{tab:prompt_variations_condensed}, \ref{tab:prompt_variations_simple}. In general, we found models' hierarchical performance across the three experiments under those prompt variations remains stable. No prompt variation shows substantial advantage over the original prompt over three experiment. The only exception was the condensed prompt improved InternVL2.5-4B-MPO and Ovis-4B's performance on Vocabulary task, while it also reduced several models' accuracy on Possessive and Demonstrative task.

\paragraph{Context-First}
\begin{quote}\itshape
``In this scene, the [speaker] is talking about their own or the other person’s preferences. The [gazer] is looking at the object in the green square, which shows the preference. Now take the role of the [speaker]. From [options], choose ONE sentence. Write only that sentence.''
\end{quote}

\paragraph{Condensed}
\begin{quote}\itshape
``Act as the [speaker]. The [speaker] discusses object preferences. The [gazer] looks at the object in the green square, showing the preference. Choose ONE sentence from [options]. Respond with only that sentence.''
\end{quote}

\paragraph{Simple}
\begin{quote}\itshape
``You are the [speaker]. The [gazer] looks at the object in the green box, which means they like it. Choose ONE sentence from [options]. Write only that sentence.''
\end{quote}

\begin{table}[t]
\centering
\small
\begin{tabular}{@{}llr@{}}
\toprule
\textbf{Experiment} & \textbf{Model} & \textbf{Accuracy} \\
\midrule
Vocabulary    & GPT-4o              & 0.999 \\
              & Llama-4-Scout       & 0.997 \\
              & Gemma-3             & 0.990 \\
              & InternVL2\_5-4B-MPO & 0.483 \\
              & InternVL2\_5-8B-MPO & 0.871 \\
              & Ovis2-4B            & 0.520 \\
              & Ovis2-8B            & 0.516 \\
\midrule
Possessive    & GPT-4o              & 0.826 \\
              & Llama-4-Scout       & 0.742 \\
              & Gemma-3             & 0.721 \\
              & InternVL2\_5-4B-MPO & 0.402 \\
              & InternVL2\_5-8B-MPO & 0.432 \\
              & Ovis2-4B            & 0.436 \\
              & Ovis2-8B            & 0.428 \\
\midrule
Demonstrative & GPT-4o              & 0.633 \\
              & Llama-4-Scout       & 0.444 \\
              & Gemma-3             & 0.480 \\
              & InternVL2\_5-4B-MPO & 0.237 \\
              & InternVL2\_5-8B-MPO & 0.319 \\
              & Ovis2-4B            & 0.236 \\
              & Ovis2-8B            & 0.242 \\
\bottomrule
\end{tabular}
\caption{Accuracy of models under the \textbf{Context-First} prompt variation.}
\label{tab:prompt_variations_context}
\end{table}

\begin{table}[t]
\centering
\small
\begin{tabular}{@{}llr@{}}
\toprule
\textbf{Experiment} & \textbf{Model} & \textbf{Accuracy} \\
\midrule
Vocabulary    & GPT-4o              & 0.996 \\
              & Llama-4-Scout       & 0.973 \\
              & Gemma-3             & 0.965 \\
              & InternVL2\_5-4B-MPO & 0.952 \\
              & InternVL2\_5-8B-MPO & 0.974 \\
              & Ovis2-4B            & 0.831 \\
              & Ovis2-8B            & 0.514 \\
\midrule
Possessive    & GPT-4o              & 0.721 \\
              & Llama-4-Scout       & 0.728 \\
              & Gemma-3             & 0.702 \\
              & InternVL2\_5-4B-MPO & 0.409 \\
              & InternVL2\_5-8B-MPO & 0.443 \\
              & Ovis2-4B            & 0.435 \\
              & Ovis2-8B            & 0.622 \\
\midrule
Demonstrative & GPT-4o              & 0.475 \\
              & Llama-4-Scout       & 0.547 \\
              & Gemma-3             & 0.490 \\
              & InternVL2\_5-4B-MPO & 0.220 \\
              & InternVL2\_5-8B-MPO & 0.243 \\
              & Ovis2-4B            & 0.262 \\
              & Ovis2-8B            & 0.246 \\
\bottomrule
\end{tabular}
\caption{Accuracy of models under the \textbf{Condensed} prompt variation.}
\label{tab:prompt_variations_condensed}
\end{table}

\begin{table}[t]
\centering
\small
\begin{tabular}{@{}llr@{}}
\toprule
\textbf{Experiment} & \textbf{Model} & \textbf{Accuracy} \\
\midrule
Vocabulary    & GPT-4o              & 0.996 \\
              & Llama-4-Scout       & 0.961 \\
              & Gemma-3             & 0.954 \\
              & InternVL2\_5-4B-MPO & 0.630 \\
              & InternVL2\_5-8B-MPO & 0.971 \\
              & Ovis2-4B            & 0.516 \\
              & Ovis2-8B            & 0.513 \\
\midrule
Possessive    & GPT-4o              & 0.672 \\
              & Llama-4-Scout       & 0.568 \\
              & Gemma-3             & 0.551 \\
              & InternVL2\_5-4B-MPO & 0.430 \\
              & InternVL2\_5-8B-MPO & 0.475 \\
              & Ovis2-4B            & 0.452 \\
              & Ovis2-8B            & 0.553 \\
\midrule
Demonstrative & GPT-4o              & 0.530 \\
              & Llama-4-Scout       & 0.376 \\
              & Gemma-3             & 0.312 \\
              & InternVL2\_5-4B-MPO & 0.247 \\
              & InternVL2\_5-8B-MPO & 0.277 \\
              & Ovis2-4B            & 0.288 \\
              & Ovis2-8B            & 0.233 \\
\bottomrule
\end{tabular}
\caption{Accuracy of models under the \textbf{Simple} prompt variation.}
\label{tab:prompt_variations_simple}
\end{table}

\clearpage
\newpage
\section{Ablation Experiment}
\label{app:ablation}

In this section, we showcase our images and prompts used for the ablation experiment. The bold text highlights the changed text in the ablated prompt and the struck-through text indicates the text was deleted in the ablated prompt.

\paragraph{Possessive Task Ablation} Spatial cue removal (linguistic only, removal of ownership information): the image used was the same as in the main Possessive experiment. The prompt used shows below:
\begin{quote}
    \textit{"You'll play the role of the [speaker] in the scene. The [speaker] is talking about [his or the girl's / her or the boy's] preferences of the objects on the table. The [gazer] is gazing at the [object] in the green square\st{, which belongs to the [owner]}. The gazing shows the preference. Your task is to select the right message as the [speaker]. IMPORTANT: Respond with ONLY ONE SENTENCE from [options]. Do not explain. Do not add any words. Just type the words."}
\end{quote}

Spatial cue removal (visual only): the prompt used was the same as in the main Possessive experiment. See figure \ref{figure-ablation-pos-visual} for the image used in this ablation.
\begin{figure}[h]
\begin{center}
\includegraphics[width=0.7\columnwidth]{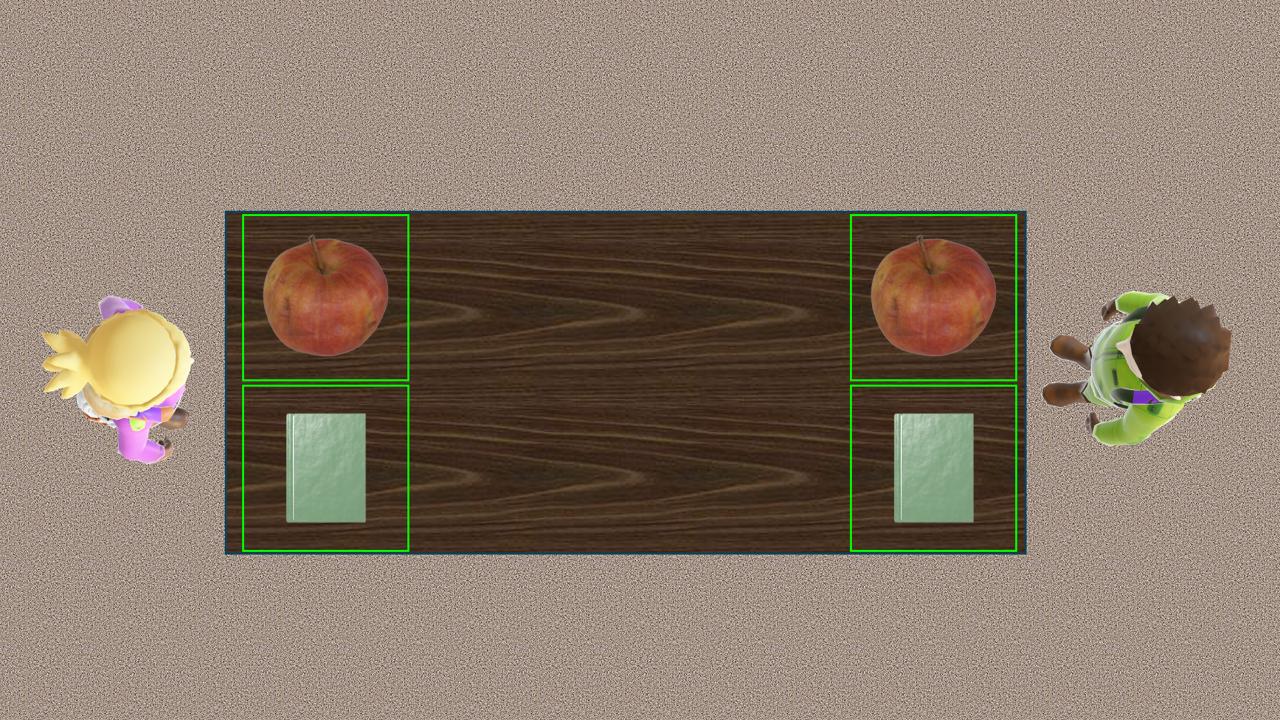}
\end{center}
\caption{Sample image in the Possessive ablation experiment, spatial cue removal (visual only). We removed the visual cue by adding green squares to all objects}
\label{figure-ablation-pos-visual}
\end{figure}

Spatial cue removal (linguistic and visual): See figure \ref{figure-ablation-pos-both} for the image used in this ablation. The prompt used shows below:
\begin{figure}[h]
\begin{center}
\includegraphics[width=0.7\columnwidth]{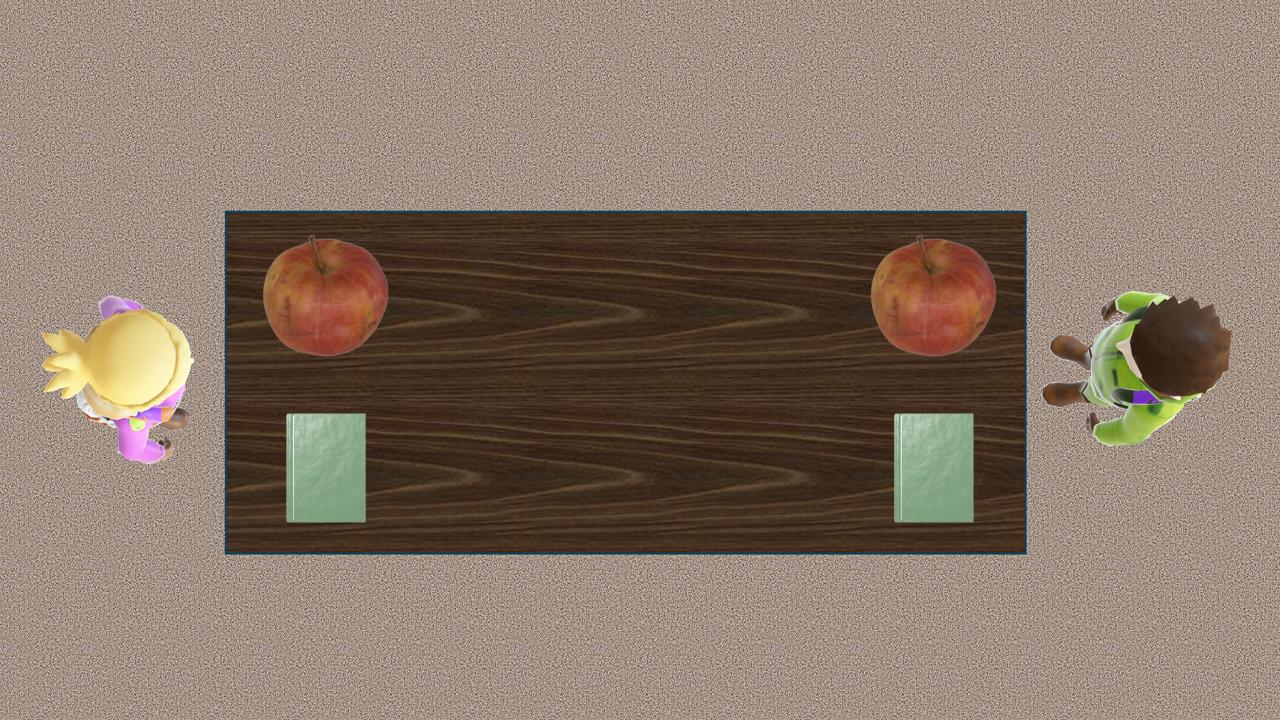}
\end{center}
\caption{Sample image in the Possessive ablation experiment, spatial cue removal (both linguistic and visual). We removed the visual cue by removing all green squares of all objects}
\label{figure-ablation-pos-both}
\end{figure}
\begin{quote}
    \textit{"You'll play the role of the [speaker] in the scene. The [speaker] is talking about [his or the girl's / her or the boy's] preferences of the objects on the table. \st{The [gazer] is gazing at the [object] in the green square, which belongs to the [owner]}. The gazing shows the preference. Your task is to select the right message as the [speaker]. IMPORTANT: Respond with ONLY ONE SENTENCE from [options]. Do not explain. Do not add any words. Just type the words."}
\end{quote}

Perspectival cue removal
The image used was the same as in the main Possessive experiment. The prompt used shows below:
\begin{quote}
    \textit{"You'll play the role of the \textbf{someone} in the scene. The [speaker] is talking about [his or the girl's / her or the boy's] preferences of the objects on the table. The [gazer] is gazing at the [object] in the green square, which belong to the [owner]. The gazing shows the preference. Your task is to select the right message\st{ as the [speaker]}. IMPORTANT: Respond with ONLY ONE SENTENCE from [options]. Do not explain. Do not add any words. Just type the words."}
\end{quote}

\paragraph{Demonstrative Task Ablation} Spatial cue removal (linguistic only, removal of object proximity information): the image used was the same as in the main Demonstrative experiment. The prompt used shows below:
\begin{quote}
    \textit{"You'll play the role of the [speaker] in the scene. The [speaker] is talking about [his or the girl's / her or the boy's] preferences of the objects on the table. The [gazer] is gazing at the [object] in the green square\st{, which is close to the [person near the target]}. The gazing shows the preference. Your task is to select the right message as the [speaker]. IMPORTANT: Respond with ONLY ONE SENTENCE from [options]. Do not explain. Do not add any words. Just type the words."}
\end{quote}

Spatial cue removal (visual only): the prompt used was the same as in the main Demonstrative experiment. See figure \ref{figure-ablation-dem-visual} for the image used in this ablation.
\begin{figure}[h]
\begin{center}
\includegraphics[width=0.7\columnwidth]{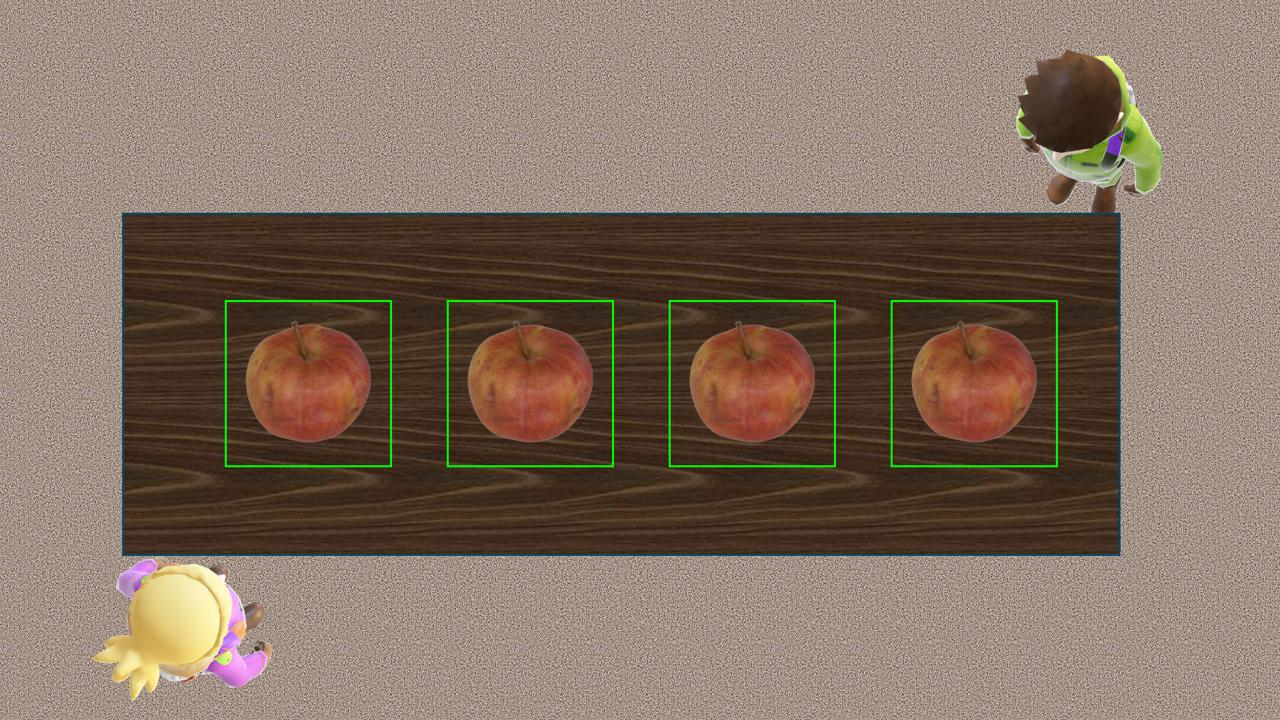}
\end{center}
\caption{Sample image in the Demonstrative ablation experiment, spatial cue removal (visual only). We removed the visual cue by adding green squares to all objects}
\label{figure-ablation-dem-visual}
\end{figure}

Spatial cue removal (linguistic and visual): See figure \ref{figure-ablation-dem-both} for the image used in this ablation. The prompt used shows below:
\begin{figure}[h]
\begin{center}
\includegraphics[width=0.7\columnwidth]{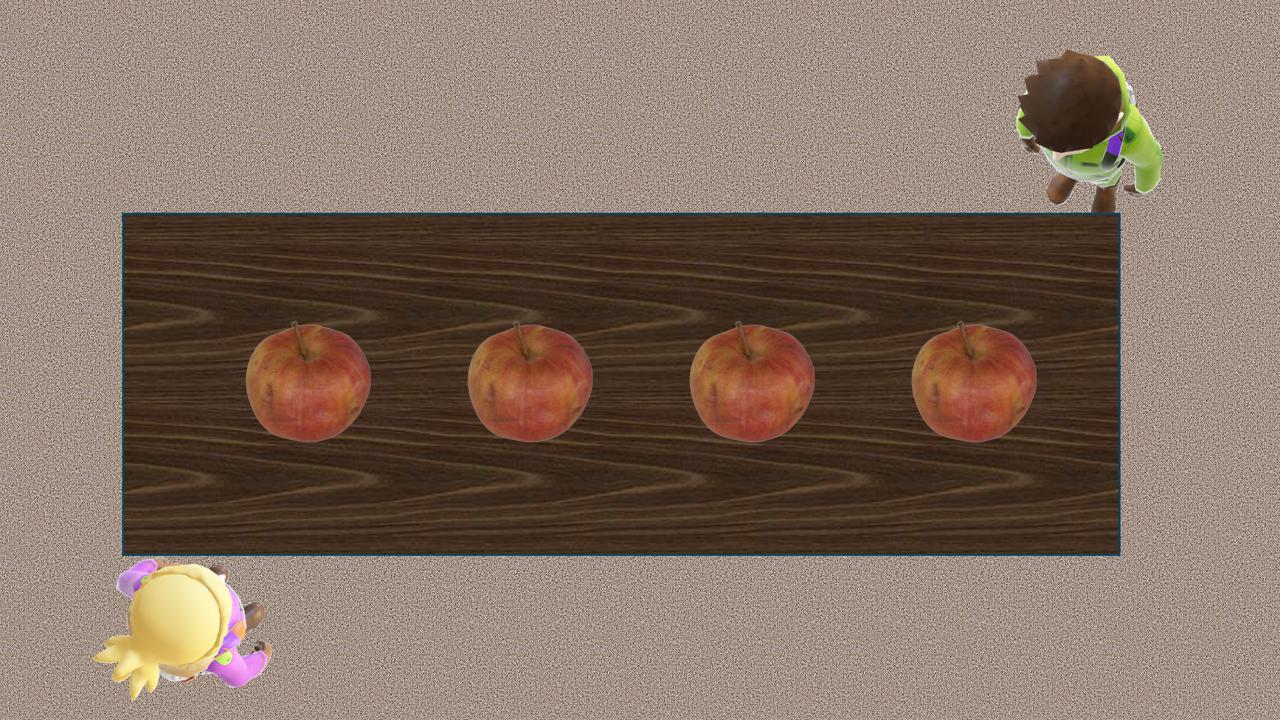}
\end{center}
\caption{Sample image in the Demonstrative ablation experiment, spatial cue removal (both linguistic and visual). We removed the visual cue by removing all green squares of all objects}
\label{figure-ablation-dem-both}
\end{figure}
\begin{quote}
    \textit{"You'll play the role of the [speaker] in the scene. The [speaker] is talking about [his or the girl's / her or the boy's] preferences of the objects on the table. \st{The [gazer] is gazing at the [object] in the green square, which is close to the [person near the target]}. The gazing shows the preference. Your task is to select the right message as the [speaker]. IMPORTANT: Respond with ONLY ONE SENTENCE from [options]. Do not explain. Do not add any words. Just type the words."}
\end{quote}

Perspectival cue removal
The image used was the same as in the main Possessive experiment. The prompt used shows below:
\begin{quote}
    \textit{"You'll play the role of \textbf{someone} in the scene. The [speaker] is talking about [his or the girl's / her or the boy's] preferences of the objects on the table. The [gazer] is gazing at the [object] in the green square, which is close to the [person near target]. The gazing shows the preference. Your task is to select the right message\st{ as the [speaker]}. IMPORTANT: Respond with ONLY ONE SENTENCE from [options]. Do not explain. Do not add any words. Just type the words."}
\end{quote}

\clearpage
\newpage
\section{Instruction-based Prompting and In-context Learning Experiments}
\label{learning}
\subsection{In-context Learning}
We first present the models four examples capturing all personal/ possessive/ demonstrative pronoun combinations with correct reference solutions. The prompt for the examples is:
\begin{quote}
    \textit{"In this picture, the [speaker] is speaking and the [gazer] is gazing at the object in the green square, and the [speaker] says '[correct reference]'"}
\end{quote}
After the example presentations, the models would receive the same image and prompt combination as in the zero-shot experiments.

\subsection{Instruction-based prompting}

We sent the prompt including task image plus hints as input to the models and obtained their answer. We use bold text to highlight the added hints to the original prompts.

\paragraph{Possessive Task}
\begin{quote}
  \textit{“You'll play the role of the [speaker] in the scene. The [speaker] is talking about [his or the girl's / her or the boy's] preferences of the objects on the table. The [gazer] is gazing at the [object] in the green square, which belongs to the [owner]. The gazing shows the preference. \textbf{Your task is to produce a reference in this format: "\{personal pronoun\} prefer \{possessive pronoun\}" as the [speaker]. Choose the correct option based on: Who is speaking (you as the [speaker]); Who is expressing preference (the gazer: the [gazer]); Who owns the object being preferred (the owner: the [owner]). Use the instructions form before to select the right message as the [speaker].} IMPORTANT: Respond with ONLY ONE SENTENCE from [options].”}
\end{quote}

\paragraph{Demonstrative Task}
\begin{quote}
  \textit{"You'll play the role of the [speaker] in the scene. The [speaker] is talking about [his or the girl's / her or the boy's] preferences of the objects on the table. The [gazer] is gazing at the [object] in the green square, which belong to the [owner]. The gazing shows the preference. \textbf{Your task is to produce a reference in this format: "\{personal pronoun\} prefer \{demonstrative pronoun\}" as the [speaker]. Choose the correct option based on: Who is speaking (you as the [speaker]); Who is expressing preference (the gazer: the [gazer]); Who is close to the object being preferred (the person being close to: the [person near target]). Use the instructions form before to select the right message as the [speaker].} IMPORTANT: Respond with ONLY ONE SENTENCE from [options]."}
\end{quote}


\clearpage
\newpage
\section{Human Comprehension Task}
\label{comprehension}
To better understand the human error patterns, a comprehension version of the Possessive and Demonstrative tasks was built where one of the two characters expressed either their own or the other's preference for an object (e.g., `I prefer yours' or `You prefer this one') and participants had to indicate on a 1-7 scale the probability that each object on the table (labeled A-D) was the referent. The Comprehension task included 8 Possessives trials and 8 Demonstratives trials, individually randomized, and was administered to a new group of native English speakers (\textit{N}=100). Figure \ref{figure-sample1VOC} and \ref{figure-sample2VOC} shows two samples from Possessive comprehension and Demonstrative comprehension tasks, respectively.

\begin{figure}[t]
\begin{center}
\includegraphics[width=0.7\columnwidth]{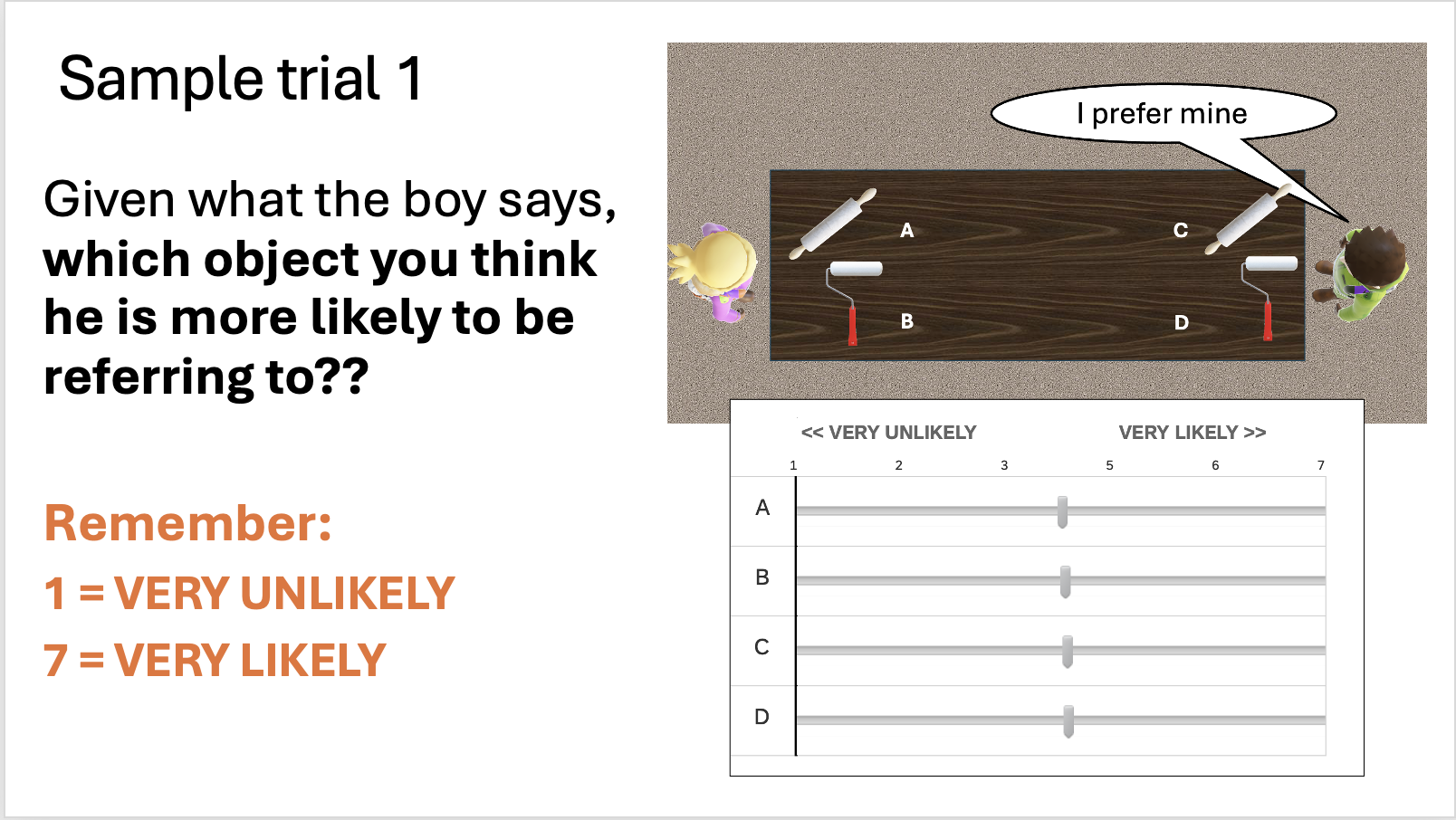}
\end{center}
\caption{Sample trial in the Possessive Comprehension task.}
\label{figure-sample1VOC}
\end{figure}

\begin{figure}[t]
\begin{center}
\includegraphics[width=0.7\columnwidth]{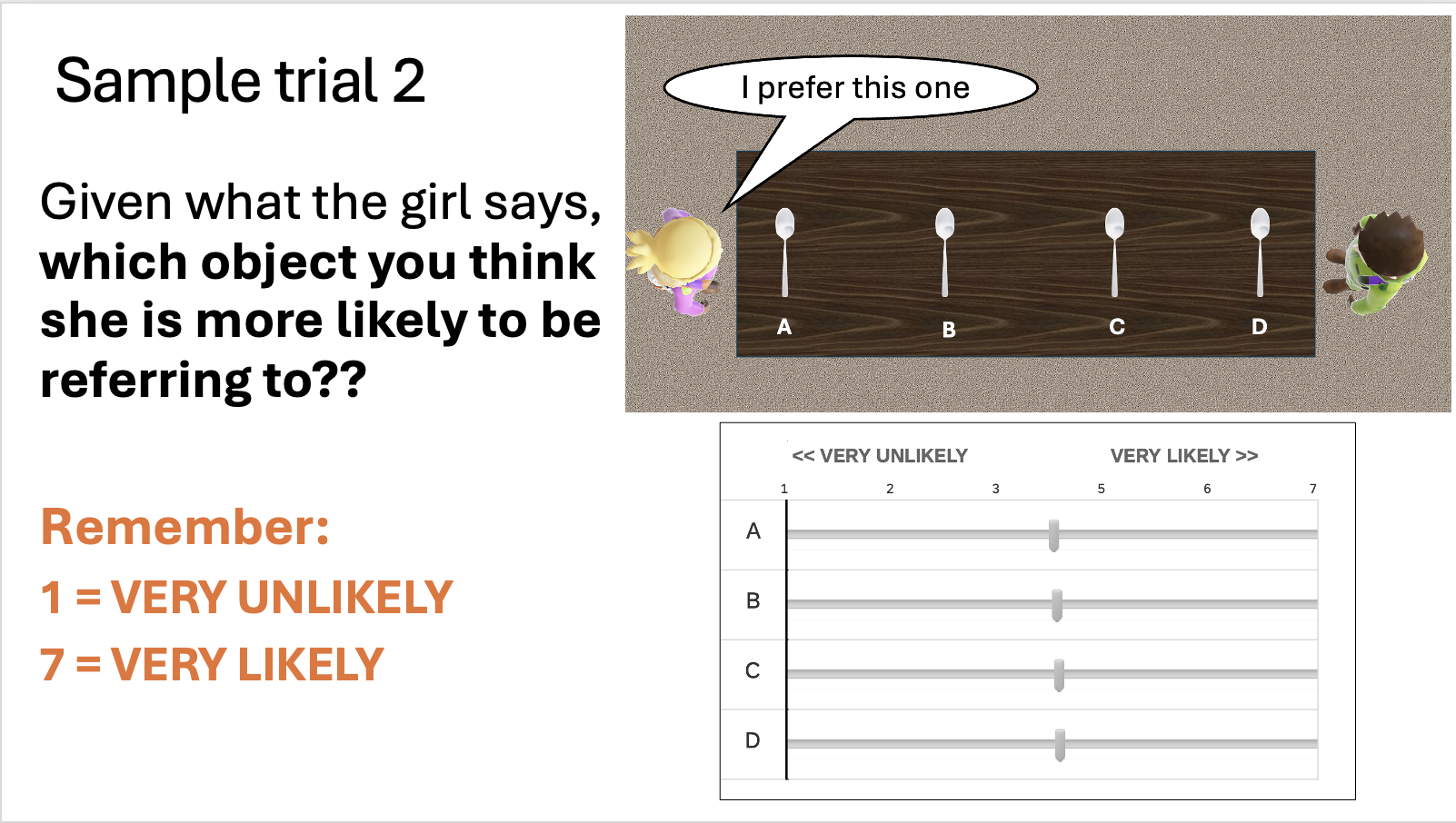}
\end{center}
\caption{Sample trial in the Demonstrative Comprehension task.}
\label{figure-sample2VOC}
\end{figure}

As the Figure \ref{figure-result_comp} shows, possessive pronouns `mine and `yours' display a clear binary distribution, with target objects receiving high likelihood ratings and non-target objects consistently being rated low. In contrast, demonstrative interpretation follows a graded pattern based on the target's distance from the speaker. For `this,' the objects closer to the speaker receive the highest likelihood ratings, while for `that,' the objects farther from the speaker are rated highest. Interestingly, we also observe subtle distinctions within demonstratives: `this' exhibits a sharper proximity-based gradient (indicating a more specific meaning), whereas `that' shows more similar ratings across intermediate and distant positions (indicating a vaguer meaning).

\begin{figure}[t]
\begin{center}
\includegraphics[width=1\columnwidth]{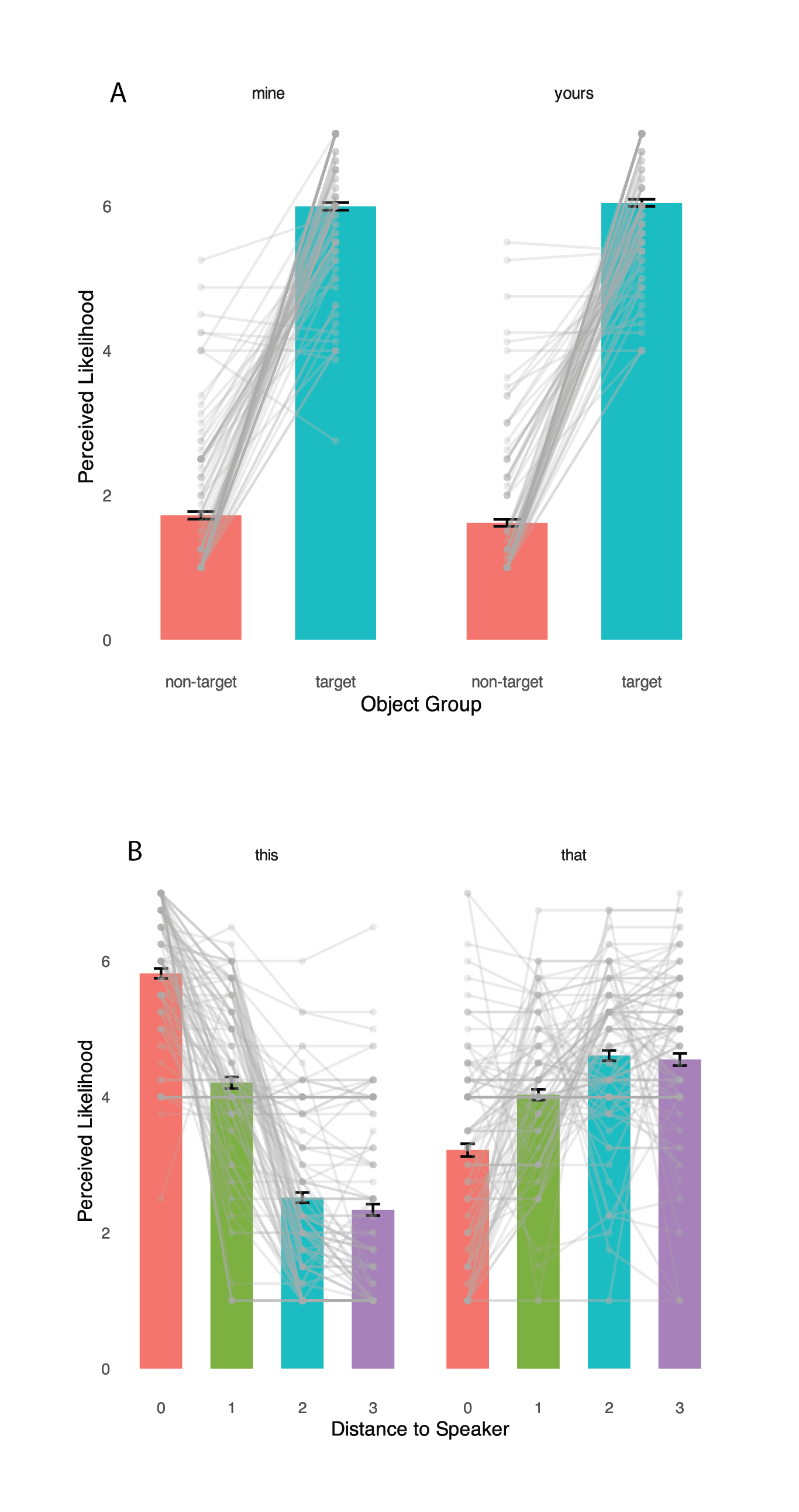}
\end{center}
\caption{\textbf{(A)} Perceived probability ratings in the Possessive comprehension task, comparing target object pair versus non-target object pair. \textbf{(B)} Perceived probability ratings in the Demonstrative comprehension task as a function of object distance from speaker, ranging from 0 (nearest) to 3 (farthest).}
\label{figure-result_comp}
\end{figure}

\clearpage
\newpage
\section{Error Pattern of Human and MLMs}
\label{error}

\begin{figure*}[h]
\centering
\includegraphics[width=1\textwidth]{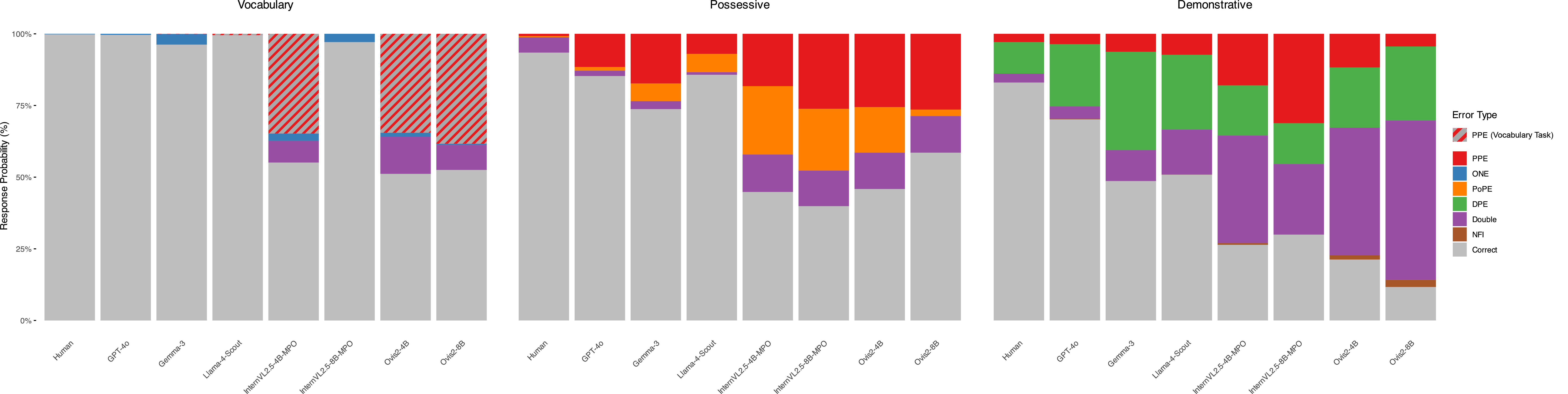}
\caption{Response probability by task type with respect to \textbf{PPE}: personal pronoun error; \textbf{ONE}: object noun error; \textbf{PoPE}: possessive pronoun error; \textbf{DPE}: demonstrative pronoun error; \textbf{Double}: both personal and possessive/demonstrative pronoun error; \textbf{NFI}: not following instructions (only in MLM evaluations), and \textbf{Correct}: correct responses. Gray lines mark PPE in the Vocabulary task, where personal pronouns were not among the response options, in contrast to the other tasks.}
\label{figure-error-pattern}
\end{figure*}

Figure \ref{figure-error-pattern} presents response distributions for humans and all models across all three tasks. For the Vocabulary task, humans and the three advanced models (GPT-4o, Llama-4-Scout, and Gemma-3) demonstrate near-perfect accuracy, while other models show errors primarily in personal pronoun selection. In the Possessive task, humans make relatively few errors, whereas even advanced models occasionally make incorrect personal pronoun choices, suggesting limitations in perspective-taking abilities that may account for their performance gap relative to humans.

Recall that the Demonstrative task presents the greatest challenge for both humans and models. Interestingly, despite quantitative differences in performance, we observe a convergence in error types between humans and the three advanced models, with the most frequent errors occurring in demonstrative pronoun use, followed by errors in personal pronoun selection or combined errors. We take this finding as evidence of the intrinsic demands that the Demonstrative task poses compared to Vocabulary and Possessive tasks to both humans and models. We do not analyze the remaining models in detail, as they perform at near-chance level and exhibit error patterns that are less interpretable.

\clearpage
\newpage
\section{Additional visual ablation baselines.}
\label{addtionalbaselines}
We conducted the additional visual ablation baselines: an MLM with an empty image, an MLM with an unrelated image, and a text-only version (no image). We evaluated all of these models on both the possessive and demonstrative tasks using the same trial structure as in our sanity check (for space reasons, we present the results for the possessive task in Table~\ref{tab:possessive}; demonstratives show a similar pattern but with overall lower accuracy). As expected, these baselines yield a larger accuracy drop than the Spatial Cue Removal (Visual) used in the paper, with especially pronounced declines for smaller models and for the unrelated-image condition. These results confirm that visual information meaningfully contributes to model performance, even though models tend to overweight linguistic cues.
\begin{table}[h]
\centering
\caption{Accuracy on the possessive task across different baseline conditions.}
\label{tab:possessive}
\begin{tabular}{lll}
\toprule
\textbf{Test Type} & \textbf{Model} & \textbf{Accuracy} \\
\midrule
no image & gpt-4o & 0.703 \\
no image & Llama-4-Scout & 0.391 \\
no image & Gemma-3 & 0.657 \\
no image & InternVL2.5-4B-MPO & 0.000 \\
no image & InternVL2.5-8B-MPO & 0.344 \\
no image & Ovis2-4B & 0.023 \\
no image & Ovis2-8B & 0.484 \\
\midrule
empty image & gpt-4o & 0.633 \\
empty image & Llama-4-Scout & 0.422 \\
empty image & Gemma-3 & 0.633 \\
empty image & InternVL2.5-4B-MPO & 0.141 \\
empty image & InternVL2.5-8B-MPO & 0.273 \\
empty image & Ovis2-4B & 0.070 \\
empty image & Ovis2-8B & 0.000 \\
\midrule
unrelated image & gpt-4o & 0.266 \\
unrelated image & Llama-4-Scout & 0.484 \\
unrelated image & Gemma-3 & 0.391 \\
unrelated image & InternVL2.5-4B-MPO & 0.070 \\
unrelated image & InternVL2.5-8B-MPO & 0.211 \\
unrelated image & Ovis2-4B & 0.312 \\
unrelated image & Ovis2-8B & 0.000 \\
\bottomrule
\end{tabular}
\end{table}

\end{document}